\documentclass[10pt,twocolumn,letterpaper]{article} 
\usepackage[accsupp]{axessibility} %
\usepackage[pagenumbers]{cvpr} %

\usepackage[dvipsnames]{xcolor}
\usepackage[table]{xcolor}
\usepackage{multirow} 
\usepackage{bbm}
\usepackage{xcolor}
\usepackage{siunitx}
\usepackage{tikz}
\usepackage{float}

\newcommand{\parnobf}[1]{\vspace{1mm} \par \noindent {\bf #1}}
\newcommand{\parnobfnovspace}[1]{\par \noindent {\bf #1}}

\newcommand{\numevalscenes}{476\xspace}

\newcommand{\vggtft}{VGGT$_\text{FT}$\xspace}
\newcommand{\picubed}{$\pi^3$\xspace}
\newcommand{\picubedft}{$\pi^3_\text{FT}$}

\newcommand{\worldmirrorft}{WM$_\text{FT}$\xspace}

\newcommand{\picubedtrans}{$\pi^3_\text{TL}$}
\newcommand{\vggttrans}{VGGT$_\text{TL}$\xspace}
\newcommand{\worldmirrortrans}{WM$_\text{TL}$\xspace}

\newcommand{\evaldataset}{MegaUnScene\xspace}

\newcommand{\cambridge}{\emph{s}ELP\xspace}
\newcommand{\wELP}{{UnScenePairs}\xspace}
\newcommand{\wELPt}{{UnScenePairs-t}\xspace}

\newcommand{\tablestyle}[2]{\setlength{\tabcolsep}{#1}\renewcommand{\arraystretch}{#2}\centering\footnotesize}

\newbox\jsavebox

 \newif\ifcomment

\definecolor{myorange}{HTML}{ff751f}
\definecolor{myblue}{HTML}{1f80ff}
\definecolor{mygreen}{HTML}{00dd1d}
\definecolor{mycyan}{HTML}{00f7ff}
\definecolor{mymagenta}{HTML}{ff1fa9}

\definecolor{cvprblue}{rgb}{0.21,0.49,0.74}
\usepackage[pagebackref,breaklinks,colorlinks,allcolors=cvprblue]{hyperref}

\title{Emergent Extreme-View Geometry in 3D Foundation Models}

\author{
Yiwen Zhang$^1$ \quad
Joseph Tung$^2$ \quad
Ruojin Cai$^3$ \quad
David Fouhey$^2$ \quad
Hadar Averbuch-Elor$^1$
\\[1mm]
$^1$Cornell University \ \ \
$^2$New York University \ \ \
$^3$Kempner Institute, Harvard University
\\[5mm]
\normalsize{\url{https://ext-3dfms.github.io/}}
}

\begin{document}

\maketitle
\begin{abstract}
{
3D foundation models (3DFMs) have recently transformed 3D vision, enabling joint prediction of depths, poses, and point maps directly from images. Yet their ability to reason under extreme, non-overlapping views remains largely unexplored. In this work, we study their internal representations and find that 3DFMs exhibit an emergent understanding of extreme-view geometry, despite never being trained for such conditions. To further enhance these capabilities, we introduce a lightweight alignment scheme that refines their internal 3D representation by tuning only a small subset of backbone bias terms, leaving all decoder heads frozen. This targeted adaptation substantially improves relative pose estimation under extreme viewpoints without degrading per-image depth or point quality. Additionally, we contribute \evaldataset, a new benchmark of Internet scenes unseen by existing 3DFMs, with dedicated test splits for both relative pose estimation and dense 3D reconstruction. All code and data will be released.

}
\end{abstract}

\section{Introduction}
\label{sec:intro}
\begin{figure}[t]
    \centering
    \includegraphics[width=0.47\textwidth]
    {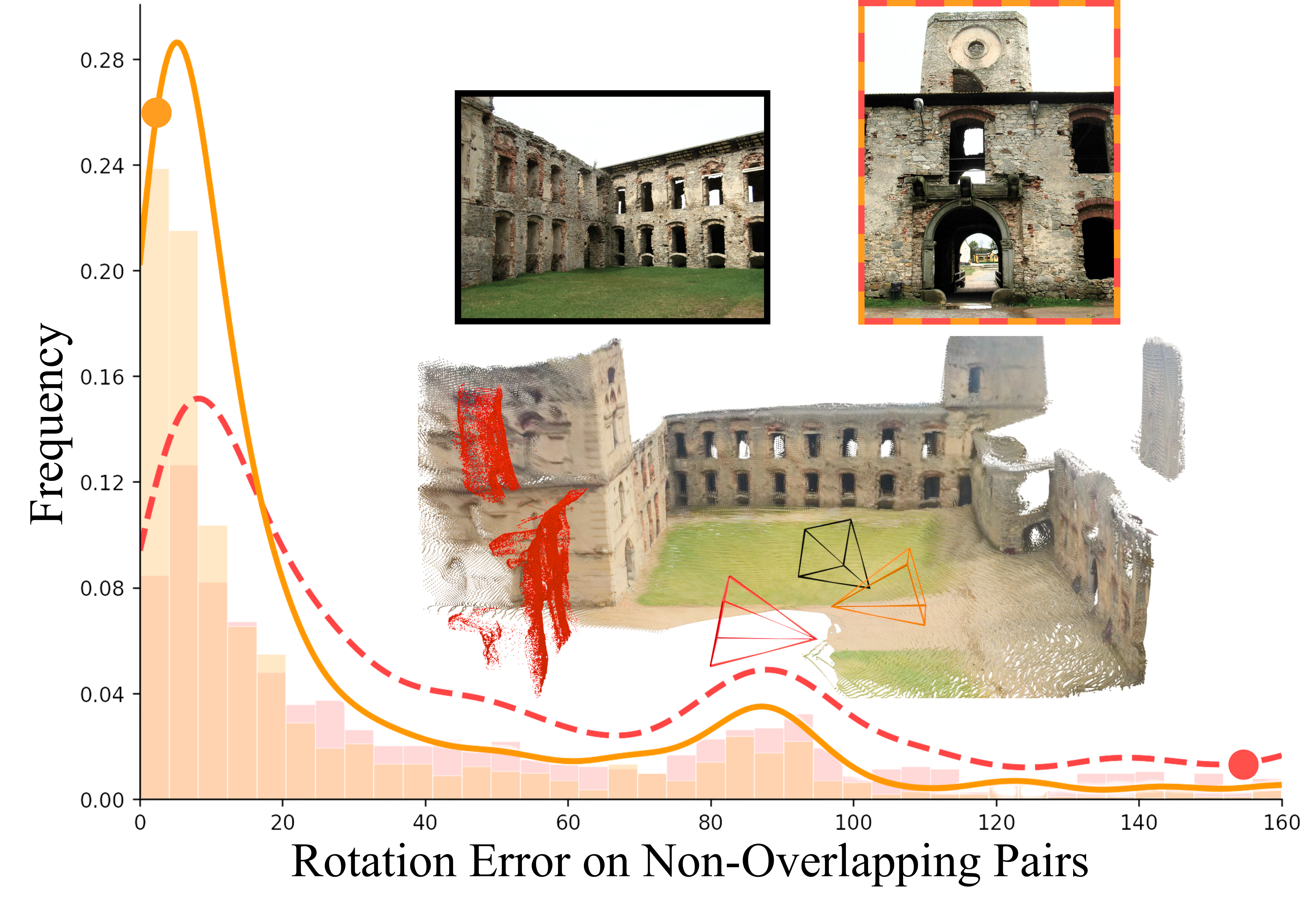}
    \vspace{-10pt}
    \caption{\textbf{Do 3D foundation models have an emergent understanding of extreme-views?}
    The \textcolor{red}{\bf pretrained} VGGT model was trained primarily on overlapping images. Surprisingly, when tested on non-overlapping image pairs$^*$, the model still produces plausible estimates of relative pose, with nearly half of the pairs yielding a rotation error below $30^\circ$.
    Careful fine-tuning of a small number of parameters \textcolor{myorange}{\bf substantially improves results}. %
    Here, for instance, the pretrained model produces incorrect pose, \textcolor{red}{\bf yielding the red ghost structure}); the  fine-tuned model corrects the error.
\\
\footnotesize{$^*$All from \emph{unseen} scenes assembled in our \evaldataset{} benchmark.  }
    }
    \label{fig:teaser}
\end{figure}

Much of 3D vision research has been built on a comfortable assumption: that we have access to densely captured, overlapping images of a scene. With such coverage, 3D structure can be inferred by matching pixels, tracking features, and triangulating points. Yet in many real-world scenarios—such as casual captures on mobile devices, historical archives or tourist photo collections—this assumption simply does not hold. Image observations may be sparse or captured from viewpoints with minimal to no visual overlap, making correspondence extraction unreliable. In these under-constrained settings, classical 3D vision pipelines begin to falter, motivating the search for models that can reason about geometry beyond the limits of visual overlap.

Recent 3D foundation models (3DFMs)~\cite{wang2025vggt,wang2025pi3,liu2025worldmirroruniversal3dworld} jointly estimate camera poses, depths, and point maps in a single feedforward pass and represent a radical departure from classical pipelines. These models consist of a shared backbone that attends within and across images, followed by heads that produce per-image 3D geometry and relative poses that place the 3D geometry in a common coordinate frame. 3DFMs eliminate the need for explicit correspondence estimation,  but are trained primarily on data that still largely conforms to the comfortable assumption of visual overlap between views. Nonetheless, when tested on views with limited overlap, these methods are surprisingly capable of reasonable predictions (see the error distribution curves in Figure~\ref{fig:teaser}), although far from their original performance.

This paper's key premise is that an internal \emph{3D language}---a learned geometric representation---crystallizes in the shared backbone of 3DFMs, with the decoder heads simply translating this language to explicit 3D outputs. 
Based on this insight, we propose a lightweight alignment scheme that targets this internal language to enhance performance on extreme views. Specifically, we \emph{freeze} the decoder heads and tune \emph{only} the biases of selected backbone layers to minimize rotation loss on camera poses. By merely updating around $80\text{k}$ parameters and training on roughly $65\text{k}$ image pairs for 2 epochs, these billion-parameter 3DFMs improve substantially in handling extreme view cases, predicting accurate relative rotations and, at the same time, per-image depth and point map estimations remain intact. As we show experimentally, many alternative approaches (\eg, unfreezing the decoder heads) lead to performance degradation across other outputs.

To evaluate our system, we contribute a new dataset named \emph{\evaldataset}. This is necessary because existing in-the-wild datasets~\cite{li2018megadepth,wu2021towers,tung2024megascenes,vuong2025aerialmegadepth} lack dedicated test splits and have been used for 3DFM training. 
\evaldataset{} comprises \numevalscenes Internet scenes \emph{unseen} by existing 3DFMs, enabling systematic evaluation of 3DFMs under realistic, unconstrained conditions for both relative pose estimation and dense 3D reconstruction.

Our contributions include: (a) a lightweight alignment scheme that targets the internal 3D representations to enable multiple 3DFMs to work well on extreme views: reducing relative rotation error while preserving their pretrained multi-task capabilities;
(b) a new benchmark composed of \numevalscenes{} Internet scenes for evaluating 3DFM methods; and (c) a new state-of-the-art in extreme view predictions that reduces median rotation error on sELP~\cite{bezalel2025extreme} (a single camera setting) from $13.2^\circ{\to}9.7^\circ$ and on in-the-wild data (with and without translations) from $42.4^\circ{\to}13.1^\circ$ and $28.4^\circ{\to}11.7^\circ$.

\section{Related Work}

\parnobfnovspace{Extreme Pose Estimation.}
Traditional relative pose estimation relies on local feature matching and geometric verification, typically assuming sufficient visual overlap between input views. Classical pipelines employ handcrafted descriptors~\cite{lowe2004distinctive,rublee2011orb} and RANSAC-based matches~\cite{fischler1981random}.
Learning-based matchers~\cite{detone2018superpoint,sarlin2020superglue,lindenberger2023lightglue,sun2021loftr,edstedt2024roma} leverage deep neural networks to learn both feature descriptors and matching, improving robustness to illumination and appearance variations. However, these overlap-dependent methods degrade under wide baselines or non-overlapping settings.

To handle sparse or overlap-free inputs, learning-based pose predictors directly infer 3D relationships from images without explicit correspondences. Diffusion-based estimators~\cite{wang2023posediffusion,zhang2024cameras} and other sparse-view or object-centric approaches~\cite{zhang2022relpose,yang2022fvor,fan2023pope,sinha2023sparsepose,lin2023relpose++,wang2023pf} demonstrate strong 3D priors but are restricted to constrained domains.
Recent works have also explored leveraging generative video priors for pose estimation~\cite{cai2025can,mao2025posecrafter}, synthesizing plausible transitions between sparse views to provide useful cues for relative pose reasoning.

At the scene level, earlier extreme-pose methods relied on geometric assumptions~\cite{chen2021wide,rockwell20228} or additional input modalities~\cite{yang2019extreme,yang2020extreme}. Several works design learning-based frameworks specifically tailored for the relative rotation task~\cite{cai2021extreme,dekel2024estimating}. Bezalel~\etal~\cite{bezalel2025extreme} recently extend these to handle uncontrolled Internet imagery. However, performance remains limited over extreme real-world image pairs.
Our work instead adapts 3D foundation models directly, fine-tuning their geometric representations to become inherently robust to extreme, non-overlapping viewpoints.

\parnobf{3D Foundation Models.}
Classical 3D reconstruction methods such as Structure-from-Motion (SfM)~\cite{agarwal2011building,hartley2003multiple,schonberger2016structure,snavely2006photo} and Multi-View Stereo (MVS)~\cite{schoenberger2016mvs,furukawa2015multi} recover camera poses and  3D structure by matching local features and jointly optimizing via bundle adjustment.
Recent feed-forward 3D models have transformed this pipeline by directly predicting camera pose, depth, point map in a single forward pass.
DUSt3R~\cite{wang2023dust3r} pioneered pairwise reconstruction by regressing dense point maps from two input images, followed by~\cite{leroy2024groundingimagematching3d,wang2025cut3r,zhang2025flare,yang2025fast3r,duisterhof2025mastrsfm}, which improve scalability and architectural efficiency.
Building on these pairwise models, VGGT~\cite{wang2025vggt} introduced a unified transformer backbone that jointly infers camera poses, depth, and point maps from multiple views. 
Subsequent models such as~\cite{wang2025pi3,keetha2025mapanythinguniversalfeedforwardmetric,liu2025worldmirroruniversal3dworld} further generalized this paradigm with permutation-equivariant reasoning and metric-scale reconstruction for large-scale scenes. Several works also extend these models for reasoning over dynamic scenes~\cite{chen2025easi3r, lin2025moviesmotionaware4ddynamic, han2025d2ust3renhancing3dreconstruction}, while others explore more efficient paradigms~\cite{yang2025fast3r,shen2025fastvggt}.

\begin{figure*}[t]
    \centering
    \includegraphics[width=\textwidth]{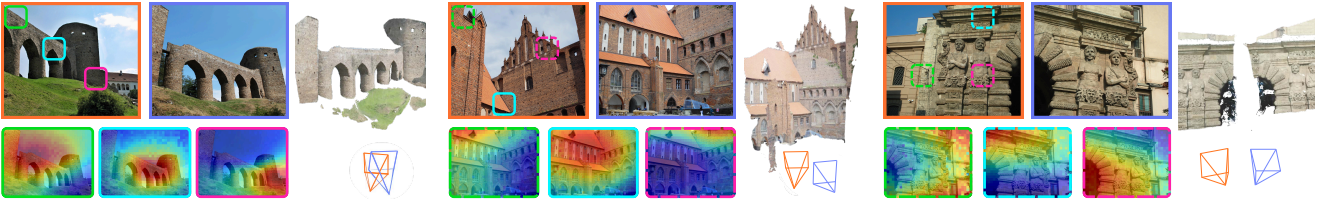}
    \vspace{-14pt}
    \caption{
        \textbf{VGGT Cross-View Attention Maps.} We visualize cross-view attention maps for three image pairs of varying overlap, from high overlap (left) to none (right). 
        For each image pair, we highlight three query regions in \textcolor{myorange}{$I_1$} with colored boxes (\textcolor{mygreen}{green}, \textcolor{mycyan}{cyan}, \textcolor{mymagenta}{magenta}). 
        Solid boxes indicate region overlap, while dashed boxes indicate no overlap. 
        \textcolor{myblue}{$I_2$}'s corresponding attention maps are shown at the bottom row with like colors. 
        Reconstructed pointmaps are shown towards the right. 
    }

    \label{fig:attention_viz}
\end{figure*}

Despite their impressive performance over diverse datasets, these foundation models are typically trained and evaluated on overlapping or smoothly varying viewpoints. 
In this work, we demonstrate that these models demonstrate an \emph{emergent} understanding of extreme-view settings, which can be further enhanced with our proposed lightweight alignment scheme that only modifies the shared alternating-attention backbone. This is unlike prior approaches~\cite{chen2025human3r, xiangli2025doppelgangersimprovedvisualdisambiguation} that typically employ a head-only fine-tuning strategy, while the shared backbone remains frozen. However, we demonstrate that this allows for significantly improving performance on extreme-view scenarios, {\it without} degrading the model's performance over other 3D tasks.

\parnobf{3D Pipeline Evaluation}. Existing benchmarks for evaluating scene-scale 3D pipelines usually rely on laser scans~\cite{schops2017multi}, controlled camera setups~\cite{jensen2014large}, or synthetic rendering~\cite{azinovic2022nrgbd}. Ideally, 3D pipelines should also handle \textit{unconstrained} inputs captured \textit{in-the-wild}. However, while datasets like MegaDepth~\cite{li2018megadepth}, WikiScenes~\cite{wu2021towers}, MegaScenes~\cite{tung2024megascenes} and AerialMegaDepth~\cite{vuong2025aerialmegadepth} use Internet photos to capture in-the-wild conditions, these datasets are typically used for training and lack dedicated evaluation subsets for dense prediction tasks. This work contributes a new evaluation benchmark assembled from unconstrained collections, enabling evaluation on unseen Internet scenes.

\section{Method}
In order to fine-tune 3D foundation models (3DFMs) for enhanced understanding of extreme geometric configurations, we propose a lightweight alignment framework that preserves the model’s pretrained knowledge while improving its robustness to large viewpoint changes. In what follows, we first analyze how 3DFMs internally represent 3D structure (Section \ref{sec:analysis}). Building upon our findings, we then introduce a compact fine-tuning scheme that aligns the shared 3D representation for viewpoint robustness (Section \ref{sec:loss}).

\subsection{The Internal Language of 3DFMs}
\label{sec:analysis}

3D foundation models (3DFMs) have recently shown remarkable progress in reconstructing scene geometry directly from unstructured images.  However, despite their growing adoption, their internal structure has remained largely unexplored. %
In this section, we first analyze their internal \emph{3D language} via cross-view attention maps, revealing that these models already encode a surprisingly rich understanding of scene geometry within their shared alternating attention backbone. We then perform a fine-grained, layer-level analysis of the backbone to examine which layers contribute the most to this learned representation.

\begin{figure}[t]
    \centering
    \includegraphics[width=\linewidth]{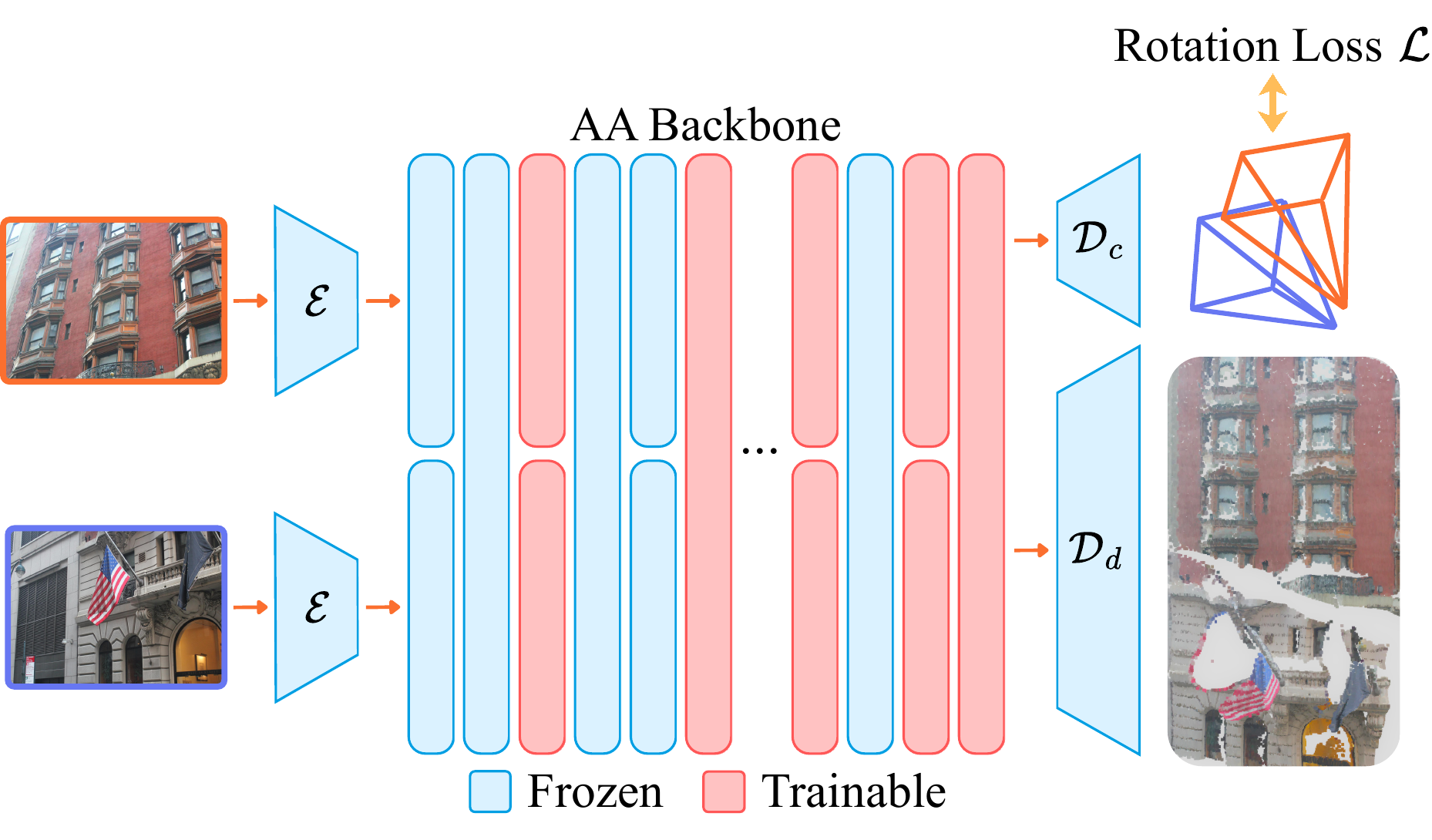}
    \vspace{-20pt}
    \caption{\textbf{Rotation-based Alignment Framework}. Above we illustrate our lightweight alignment scheme, which supervises the camera head of 3DFMs via a rotation loss $\mathcal{L}$ over predicted and ground truth relative rotation matrices. %
    To preserve pretrained knowledge, we only update the bias terms of a sparse set of layers in the shared alternating attention (AA) backbone.
    }
    \label{fig:architectural_design}
\end{figure}

\parnobf{Background: 3DFM Architectural Design.} 
Modern 3DFMs~\cite{wang2025vggt,wang2025pi3,liu2025worldmirroruniversal3dworld} share a common architectural structure, as illustrated in Figure~\ref{fig:architectural_design}. %
Each model first encodes the input images, extracting patch-level embeddings (optionally concatenated with additional input modality tokens) via an encoder $\varepsilon$. For simplicity, we describe a scenario with two input images, $I_1$ and $I_2$, each divided into $N_p = H_p \times W_p$ patches. These $N_p$ patch tokens and then fed into a shared transformer-based backbone, which alternates between \emph{frame} attention blocks and \emph{global} attention blocks. %
In a frame attention block, patch tokens are flattened and processed independently as:
\begin{equation}
\mathbf{T}_{\text{frame}}^{(i)} \in \mathbb{R}^{N_p \times D}, \quad i \in \{1,2\}.
\end{equation}
By contrast, the global attention block concatenates tokens from both images:
\begin{equation}
\mathbf{T}_{\text{global}} = [\mathbf{T}_{\text{frame}}^{(1)}, \mathbf{T}_{\text{frame}}^{(2)}] \in \mathbb{R}^{(2N_p) \times D}.
\end{equation}
Both frame and global attention use self-attention, which projects these concatenated tokens into queries $\mathbf{Q}=f_Q(\mathbf{T})$, keys $\mathbf{K}=f_K(\mathbf{T})$, and values $\mathbf{V}=f_V(\mathbf{T})$ through learned linear projections $f_Q, f_K, f_V$. For each layer $l$ and head $h$, the attention maps are computed as:
$\mathbf{A}_h^{(l)} = \text{softmax}(\mathbf{Q} \mathbf{K}^T / \sqrt{d_h})$
where $d_h$ is the dimension of the attention head $h$.

The output patch tokens are then decoded by task-specific heads, such as the camera head $\mathcal{D}_c$ and the dense prediction head $\mathcal{D}_d$ visualized in Figure \ref{fig:architectural_design}.

\parnobf{Cross-View Attention}. To understand the cross-view dependencies in the alternating attention backbone, we analyze the global attention maps, considering the last layer where the cross-view information is thoroughly fused. Specifically, we select query vectors $\mathbf{q}\in \mathbf{Q}$ located at various spatial locations in $I_1$, comparing it with all key vectors $\mathbf{k} \in \mathbf{K}$ in $I_2$ and summing over all the attention heads. 

We illustrate the cross-view attention maps in Figure \ref{fig:attention_viz}, as exemplified by VGGT~\cite{wang2025vggt} and three image pairs with varying levels of overlap. For each image pair, we highlight attention maps from three query locations, marked by green, cyan and magenta boxes. As shown, consistent patterns are observed across both \emph{overlapping} and \emph{non-overlapping} image regions. 
For regions with direct visual overlap (visualized with solid boxes), the high attention areas in $I_2$ accumulate precisely at the corresponding locations. This demonstrates the model's ability to identify exact visual correspondences  within the shared attention backbone. For regions without direct visual overlap (visualized with dashed boxes), the cross-view maps reveal non-trivial structural patterns. For instance, in the middle example, regions green and magenta are not visible in $I_2$, yet the attention weights accumulate at the image corners that are spatially nearest to the selected regions in the 3D world. Such ``near-correspondence'' understanding can also be observed for the image pair on the right, which contain no overlapping regions. Furthermore, for query regions that do not have a near-corresponding area (\emph{e.g.}, the rightmost green query), we can see that the model has learned to attend to other symmetry-based cues, such as the gate's overall curvature and its fine-grained ornate details, which can further guide the model's 3D understanding in such extreme cases. %

These observations suggest that a rich internal representation of the scene’s geometry is already constructed within the shared backbone. This representation---which extends far beyond direct visual correspondences---provides an implicit \emph{3D language} that encodes spatial relations, depth, and pose. The task-specific heads, such as the DPT head and camera head, simply act as format converters that read out this shared state into explicit geometric quantities like depth maps, point maps, or camera poses. 

\parnobf{Are All Backbone Layers Equally Important?} Prior studies have shown that not all layers in Transformer-based architectures contribute equally to the learned representation~\cite{bhojanapalli2021leveragingredundancyattentionreuse, he2024matterstransformersattentionneeded, jiang2025tracingrepresentationprogressionanalyzing}. Motivated by these findings, we conduct a similar analysis over 3DFMs, quantifying the degree of representational change between neighboring backbone layers. Our results reveal that only a small subset of layers exhibit substantial changes in their feature representations. Interestingly, the layers identified by this analysis largely coincide with those connected to the dense prediction head, implying that such skip connections play a key role in shaping the model’s internal 3D language; additional experimental details are provided in the supplementary material.

\subsection{Rotation-Based 3DFM Alignment}
\label{sec:loss}

The analysis we conducted in the previous section suggests that 3DFMs already encode a rich 3D language within their shared attention backbone. Building on this insight, we demonstrate that the model's representation can be significantly strengthened by supervising only on camera pose. Crucially, rather than retraining the task-specific heads, we apply minimal fine-tuning to the alternating attention layers, aligning the model's internal 3D language while avoiding overfitting to individual tasks. An overview of our approach is provided in Figure \ref{fig:architectural_design}.

\parnobf{Rotation-Only Supervision.} Inspired by prior work that design frameworks for estimating relative rotations between non-overlapping image pairs~\cite{cai2021extreme,dekel2024estimating,bezalel2025extreme}, we propose a rotation-based alignment objective over image pairs. Formally, given an image pair $(I_1, I_2)$ with predicted absolute rotation matrices $\mathbf{R}_1^{\mathrm{pred}}$ and $\mathbf{R}_2^{\mathrm{pred}}$, and ground-truth rotations $\mathbf{R}_1^{\mathrm{gt}}$ and $\mathbf{R}_2^{\mathrm{gt}}$, we propose a simple unified objective shared across all architectures:
\begin{equation}
\begin{aligned}
    \mathcal{L} 
        &= \mathcal{L}_{\mathrm{geo}}\!\left(
            \mathbf{R}_{\mathrm{rel}}^{\mathrm{pred}},
            \mathbf{R}_{\mathrm{rel}}^{\mathrm{gt}}
        \right)
        + \mathbb{1}_{\mathrm{a}}\,
           \mathcal{L}_{\mathrm{geo}}\!\left(
               \mathbf{R}_{1}^{\mathrm{pred}},
               \mathbf{I}
           \right),
\end{aligned}
\label{eq:geo_loss}
\end{equation}
where the relative rotations are 
$\mathbf{R}_{\mathrm{rel}}^{\mathrm{pred}}
  = \mathbf{R}_{2}^{\mathrm{pred}}
    (\mathbf{R}_{1}^{\mathrm{pred}})^{\top}$ and 
$\mathbf{R}_{\mathrm{rel}}^{\mathrm{gt}}
  = \mathbf{R}_{2}^{\mathrm{gt}}
    ( \mathbf{R}_{1}^{\mathrm{gt}} )^{\top}
$
and $\mathcal{L}_{\mathrm{geo}}$ denotes the geodesic loss which measures the minimal angular distance between rotation matrices on the $\mathrm{SO}(3)$ manifold. 
The indicator $\mathbb{1}_{\text{a}}$ activates the optional anchoring term for models assuming a fixed reference frame (e.g., VGGT), enforcing the first image to align with the world coordinate frame. 
For permutation-invariant architectures (e.g., \picubed), $\mathbb{1}_{\text{a}} = 0$, and the loss reduces to the symmetric relative-rotation term.

Note that, unlike the dense point-wise annotations required for training these models, our rotation-only supervision relies on much sparser signals (\emph{i.e.}, relative rotations between image pairs).

\parnobf{Preserving Pretrained Knowledge via Minimal Backbone Fine-Tuning.} The proposed rotation-based objective supervises only the camera head, without directly affecting the dense prediction head. This partial supervision can disrupt the geometric alignment between heads, degrading performance on downstream tasks that rely on dense predictions (which we show experimentally). To prevent this, we adopt a minimal backbone fine-tuning strategy that targets both the minimal set of \emph{layers} and \emph{parameters} within the    backbone: we fine-tune only the bias terms within the sparse set of backbone layers identified in Section~\ref{sec:analysis}. Prior work has shown that bias-only fine-tuning in Transformer-based models is often competitive with full fine-tuning~\cite{zaken2022bitfitsimpleparameterefficientfinetuning}. As our experiments show, combining selective layer updates with bias-only tuning allows for effectively aligning the model’s internal 3D language for extreme-view reasoning, all while preserving its pretrained multi-task knowledge. %

\parnobf{Training Data.} We follow the methodology introduced in Bezalel \etal~\cite{bezalel2025extreme} to construct a training set of image pairs from scene-level COLMAP reconstructions from MegaScenes~\cite{tung2024megascenes}. We include pairs with larger translations than~\cite{bezalel2025extreme} in order to train a model that is able to generalize well to pairs with camera translations.
We report precise details in the supplementary material.

\section{The \evaldataset Benchmark}
\label{sec:dataset}

Existing benchmarks evaluating 3DFMs typically have scenes with constrained 3D environments---\emph{e.g.}, assuming constant illumination, transient objects, and camera intrinsics. %
To evaluate 3DFMs on unconstrained inputs captured \textit{in-the-wild}, we create \emph{\evaldataset}: a new collection of \numevalscenes Internet scenes \textit{unseen} by existing models. %
From these scenes, we assemble two test sets for relative pose estimation 
and one for dense reconstruction.

\parnobf{Benchmark Construction.}
We follow the protocol from MegaScenes~\cite{tung2024megascenes} to curate a 3D reconstruction benchmark of Internet photos. We only include scenes that are not included in MegaScenes~\cite{tung2024megascenes}, verified by cross-referencing unique image and scene names.
To improve reconstruction fidelity, we slightly modify the pipeline by using Doppelgangers++~\cite{xiangli2025doppelgangersimprovedvisualdisambiguation} integrated with MASt3R-SfM~\cite{duisterhof2025mastrsfm}. Doppelgangers++ disambiguates challenging internet photos depicting ambiguous views, while dense MASt3R~\cite{leyor2024mast3r} matches yield more robust pairwise poses for incremental SfM. Lastly, we run multi-view stereo to obtain dense depth.
Additional details are in the supplementary material.

\parnobf{Benchmarking Relative Pose in the Wild.}
We construct subsets with and without camera translations for evaluating relative pose estimation: \emph{\wELP{}}, contains image pairs predominant rotational motion, and \emph{\wELPt} contains image pairs with larger camera baselines. Unlike the ELP test sets~\cite{bezalel2025extreme}, these subsets capture unconstrained, in-the-wild image pairs unseen by 3DFMs. We follow the image–pair selection procedure proposed in prior work~\cite{bezalel2025extreme}, constructing mutual $K$-nearest neighbor graphs using camera translation distances. %
We select a small $K$ for constructing \wELP{} (setting $K=5$), and a much larger value to admit pairs with greater translations for constructing \wELPt (setting $K=50$). After obtaining the candidate pairs, we assign each pair to an overlap category (Large, Small, None) according to the relative rotation between the two camera poses and their FoV.

For \wELPt, the increased translation baseline introduces parallax challenges where the classification algorithm may classify image pairs with some overlap into the \textit{None} category. Therefore, to ensure reliable overlap labels under larger baselines, we verify correspondences by querying the filtered image matches from MASt3R-SfM~\cite{duisterhof2025mastrsfm} and Doppelgangers++~\cite{xiangli2025doppelgangersimprovedvisualdisambiguation}, confirming no matches for \textit{None} pairs and distinguishing \textit{Large} from \textit{Small} overlap based on overall spatial match coverage.

Finally, we manually review all selected pairs and removed those affected by motion blur, significant occlusions, or insufficient geometric structure (e.g., flat painted surfaces). The final \wELP{} test set has 3,883 (778 non-overlapping) image pairs across 458 scenes and the \wELPt{} test set has 2,432 (763 non-overlapping) image pairs across 387 scenes. %

\parnobf{Benchmarking Dense Predictions in the Wild.}
\begin{figure}[t]
    \centering
    \includegraphics[width=0.47\textwidth]{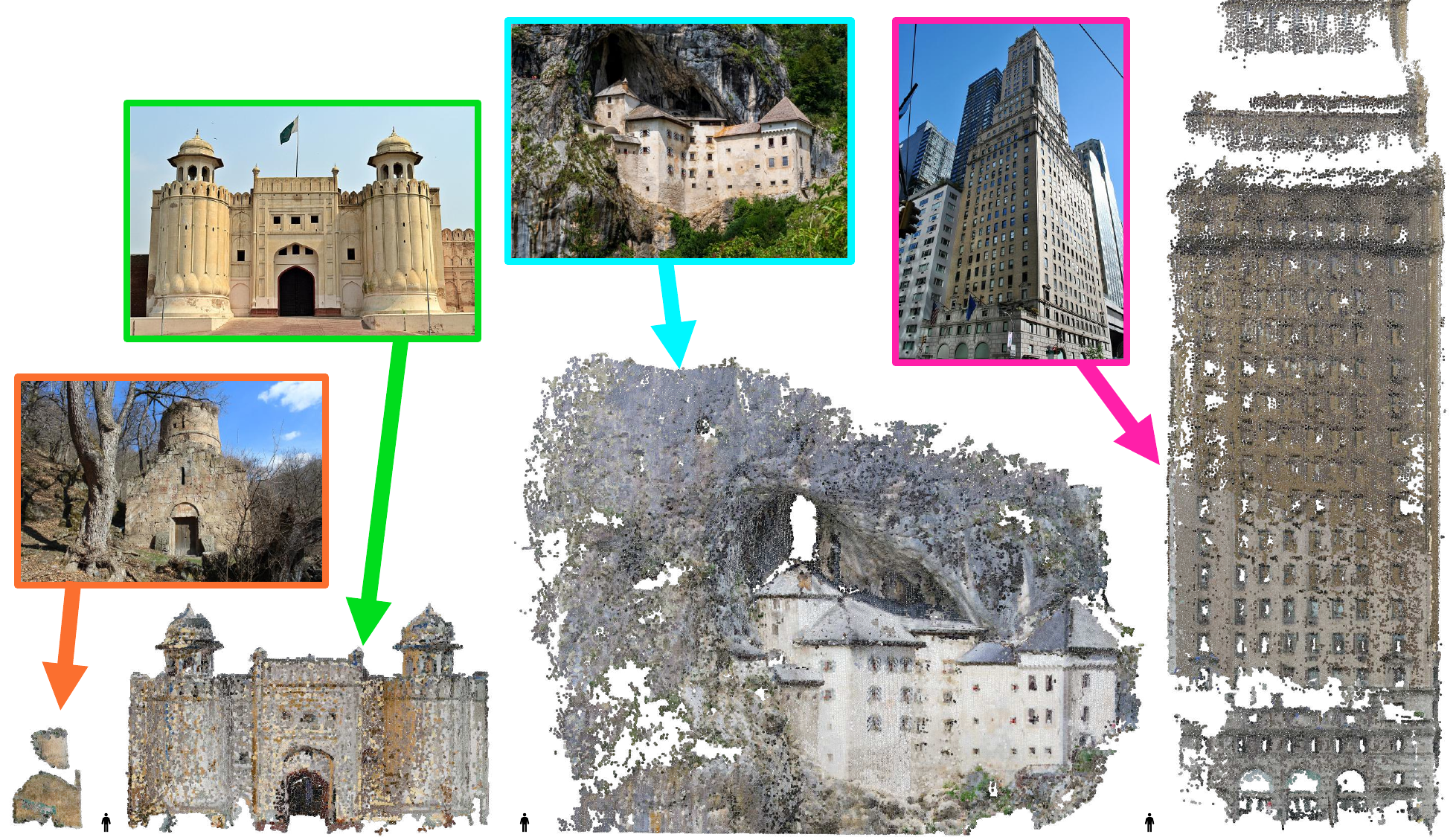}
    \vspace{-8pt}
    \caption{
    Metric scale visualization of \emph{UnSceneRecon} scenes (L→R): Aghavnavank Monastery, Alamgiri Gate, Predjama Castle, and the Ritz Tower. For reference, the person is 2 meters tall.
    }
    \label{fig:unscenerecon_vis}
\end{figure}

We construct \textit{UnSceneRecon}, a dense reconstruction subset comprising 100 in-the-wild reconstructions with metric scale annotations (Figure~\ref{fig:unscenerecon_vis}).
To curate this subset, we select reconstructions with at least 50 images and fuse depth maps to create a representative point cloud. Human annotators inspect a subset of these reconstructions, focusing on ones with the most registered images, and filter out incorrect or low-quality reconstructions. To enable comparable evaluation metrics across scenes, human annotators also identify a metric scale factor for each reconstruction by cross-referencing distances on Google Maps. Please refer to the supplementary material for additional details.

\section{Experiments}
\label{sec:results}
In this section, we present both qualitative and quantitative experiments to evaluate the effectiveness of our alignment scheme. We first conduct experiments on extreme relative rotation estimation (Section \ref{sec:rot_eval}). We then examine whether pretrained knowledge is preserved via multiview pose estimation (Section \ref{sec:gencam_eval}) and dense reconstruction (Section \ref{sec:recon_eval}). Finally, we present an ablation study comparing alternative fine-tuning strategies (Section \ref{sec:ablation}). 
Implementation details, extended ablations, additional tasks, and interactive visualizations are provided in the supplementary material. 

\parnobf{Models Considered}. We consider three recent 3DFMs:  VGGT~\cite{wang2025vggt}, WorldMirror (WM)~\cite{liu2025worldmirroruniversal3dworld} and \picubed~\cite{wang2025pi3}. Fine-tuned variants are denoted by the subscript $\text{FT}$. %

\begin{table}[t]
    \centering
    \caption{\textbf{Extreme Relative Rotation Estimation}. We evaluate performance over non-overlapping image pairs in \cambridge{}~\cite{bezalel2025extreme}, \wELP{}, and \wELPt{}. As illustrated, our fine-tuned models consistently improve the pretrained 3DFMs.%
    }
    \vspace{-0.75em}
    \tablestyle{1pt}{1.05}
    \resizebox{1.0\columnwidth}!{
    \begin{tabular}{lccccccccc}
        \toprule
        {\multirow{3}{*}{\textbf{Method}}} &
        \multicolumn{3}{c}{\textbf{\cambridge{}}} &
        \multicolumn{3}{c}{\textbf{\wELP{}}} &
        \multicolumn{3}{c}{\textbf{\wELPt{}}} \\
        \cmidrule(r){2-4} \cmidrule(r){5-7} \cmidrule(r){8-10}
        &
        {\fontsize{7pt}{7pt}\selectfont MRE}$\downarrow$ &
        {\fontsize{7pt}{7pt}\selectfont RA$_{15}$}$\uparrow$ &
        {\fontsize{7pt}{7pt}\selectfont RA$_{30}$}$\uparrow$ &
        
        {\fontsize{7pt}{7pt}\selectfont MRE}$\downarrow$ &
        {\fontsize{7pt}{7pt}\selectfont RA$_{15}$}$\uparrow$ &
        {\fontsize{7pt}{7pt}\selectfont RA$_{30}$}$\uparrow$ &
        
        {\fontsize{7pt}{7pt}\selectfont MRE}$\downarrow$ &
        {\fontsize{7pt}{7pt}\selectfont RA$_{15}$}$\uparrow$ &
        {\fontsize{7pt}{7pt}\selectfont RA$_{30}$}$\uparrow$ \\

        \midrule
        
        ExRot
        & 13.23 & 53.1 & 59.6 
        & 28.48 & 35.7 & 50.8
        & 42.45 & 31.3 & 43.8 \\
        
        \midrule

        VGGT%
        & 92.92 & 24.2 & 29.1
        & 31.64 & 33.8 & 48.8
        & 46.65 & 29.1 & 42.1 \\
        \vggtft
        & \textbf{14.21} & \textbf{50.9} & \textbf{56.6}
        & \textbf{12.71} & \textbf{53.6} & \textbf{67.9}
        & \textbf{14.48} & \textbf{50.6} & \textbf{62.1} \\

        \midrule

        WM%
        & 68.96 & 36.3 & 42.5
        & 19.25 & 44.1 & 58.9
        & 21.52 & 42.6 & 57.4 \\
        \worldmirrorft
        & \textbf{9.74} & \textbf{56.9} & \textbf{63.5}
        & \textbf{11.75} & \textbf{56.2} & \textbf{68.1}
        & \textbf{13.13} & \textbf{53.3} & \textbf{64.5} \\

        \midrule

        \picubed%
        & 45.24 & 43.8 & 48.3
        & 17.66 & 46.5 & 59.4
        & 21.62 & 43.5 & 56.8 \\
        \picubedft
        & \textbf{11.96} & \textbf{53.7} & \textbf{60.0}
        & \textbf{12.92} & \textbf{54.0} & \textbf{69.2}
        & \textbf{13.31} & \textbf{53.1} & \textbf{65.5} \\

        \bottomrule
    \end{tabular}
    }
    \label{tab:main_comparison}
\end{table}

\definecolor{myred}{HTML}{ff5250}
\definecolor{myblue}{HTML}{2563ca}
\definecolor{myyellow}{HTML}{fbb52a}
\begin{figure*}[t]
    \centering
    \includegraphics[width=0.98\textwidth]{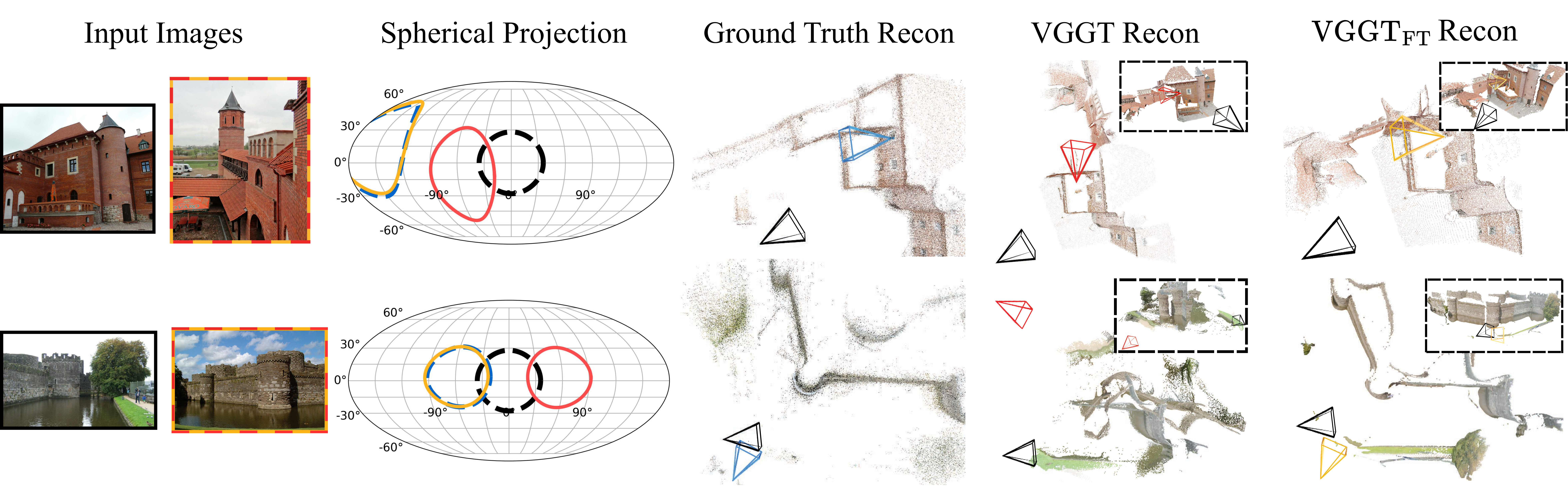}
    \vspace{-8pt}
    \caption{\textbf{Qualitative results over \wELPt{}}.
    From left to right, we show the input image pair, spherical projection of relative rotations
    (\textcolor{black}{black}: reference view, \textcolor{myblue}{blue}: ground truth,
    \textcolor{myred}{red}: pretrained VGGT, \textcolor{myyellow}{yellow}: fine-tuned VGGT),
    and the corresponding reconstructions (sparse ground truth, dense pretrained and fine-tuned). Please refer to the supplementary material for many additional visualizations. %
    }
    \label{fig:grid}
\end{figure*}

\subsection{Extreme Relative Rotation Estimation}
\label{sec:rot_eval}
\noindent \textbf{Experimental Details.} We evaluate performance on sELP~\cite{bezalel2025extreme} and the in-the-wild sets introduced in the previous sections (\wELP{}, \wELPt{}). sELP is curated from the Cambridge Landmarks dataset~\cite{kendall2015cambridgelandmarks}.  %
Beyond 3DFMs, we also compare against ExRot~\cite{bezalel2025extreme}, a recent framework tailored for the extreme rotation task. ExRot was previously state-of-the-art over in-the-wild image pairs.

\parnobf{Metrics.}
Let $\mathbf{R}_{\mathrm{pred}}$ and $\mathbf{R}_{\mathrm{gt}}$ denote the predicted and ground-truth relative rotation matrix. 
We compute the geodesic error, defined as %
$\textrm{arccos}(\frac{1}{2} (\textrm{tr}(\mathbf{R}_\textrm{pred}^\top \mathbf{R}_\textrm{gt}) - 1))$.
Following prior work~\cite{bezalel2025extreme}, we report the median rotation error (MRE) and relative rotation accuracy (RA) at thresholds of $15^\circ$ and $30^\circ$, denoted $\mathrm{RA}_{15}$ and $\mathrm{RA}_{30}$.

\parnobf{Evaluation.} 
Results over non-overlapping pairs are reported in 
Table~\ref{tab:main_comparison}. As shown, our fine-tuned models achieve consistent and substantial improvements across all test sets, establishing a new state of the art in extreme rotation estimation.
Interestingly, pretrained models show pronounced variability across test sets, performing far better on \wELP{} and \wELPt{} than on sELP{} (e.g., VGGT: MRE 31.64/46.65 vs. 92.92). This disparity likely stems from their pretraining on landmark-centric data (specifically MegaDepth~\cite{li2018megadepth}), which does not resemble the forward-walking, low-parallax trajectories characteristic of the non-overlapping sELP set. In contrast, our fine-tuned models effectively close this gap, despite using only landmark-centric data, demonstrating improved generalization.
This generalization also extends to image pairs with translation (\wELPt{}), demonstrating robustness beyond purely rotational motion. 
Figure~\ref{fig:grid} illustrates this further, showing camera pose and dense reconstruction for two pairs from \wELPt{}. After alignment, the fine-tuned VGGT predicts both accurate relative pose and translation, correcting the significant errors of the pretrained model.

Results on pairs with Large and Small overlap are reported in the supplementary material. Overall, we observe that pretrained and fine-tuned models perform similarly for overlapping pairs, with the fine-tuned variants showing consistent, moderate gains. These findings further show that our alignment procedure improves robustness in extreme-view geometry settings while preserving strong performance in overlapping cases.

\parnobf{Beyond Relative Rotations.} In the supplementary material, we also report translation accuracy on \wELPt{}. We find that while fine-tuning substantially improves rotation estimation, translation accuracy improvements are more modest. For example, the median translation error (reported in prior work~\cite{wang2024posediffusionsolvingposeestimation}) slightly decreases from 37.28$^\circ$ to 35.79$^\circ$ after fine-tuning VGGT.
We also observe the same behavior when supervising full poses rather than rotations exclusively. These results suggest that large translational displacements remain intrinsically challenging and represent an important direction for future work.

\begin{table}[t]
    \centering
    \caption{\textbf{Multiview Pose Estimation}. We evaluate camera pose angular accuracy over image collections from  RealEstate10K~\cite{zhou2024stereomagnif} and ETH3D~\cite{schops2017multi}. Non-negligible differences ($>1$ absolute) between the base and fine-tuned models are in \textbf{bold}.}
    \vspace{-0.75em}
    \tablestyle{3pt}{1.05}
    \resizebox{1.0\columnwidth}!{
    \begin{tabular}{lcccccc}
        \toprule
        \multicolumn{1}{l}{\multirow{3}{*}{\textbf{Method}}} &
        \multicolumn{3}{c}{\textbf{RealEstate10K}} &
        \multicolumn{3}{c}{\textbf{ETH3D}} \\
        \cmidrule(r){2-4} \cmidrule(r){5-7}
        \multicolumn{1}{c}{} &
        $\text{RA}_{30}\!\uparrow$ & $\text{TA}_{30}\!\uparrow$ & $\text{AUC}_{30}\!\uparrow$ &
        $\text{RA}_{30}\!\uparrow$ & $\text{TA}_{30}\!\uparrow$ & $\text{AUC}_{30}\!\uparrow$ \\
        \midrule

VGGT & 99.91 & 92.75 & 77.28 & 96.41 & 87.01 & 72.52 \\
\vggtft & 99.99 & \textbf{93.77} & \textbf{79.11} & 96.41 & \textbf{94.70} & \textbf{79.30} \\

        \midrule

WM & 99.99 & 95.42 & 85.93 & 92.48 & \textbf{91.45} & \textbf{78.24} \\
\worldmirrorft & 99.99 & 95.72 & 85.52 & 92.48 & 90.26 & 74.95 \\
        
        \midrule

\picubed & 99.99 & 95.52 & \textbf{87.14} & 100.00 & 95.38 & \textbf{82.81} \\
\picubedft & 99.99 & 95.21 & 85.25 & 100.00 & 95.21 & 79.81 \\
                
        \bottomrule
    \end{tabular}
    }
    \label{tab:relpose-angular}
\end{table}

\subsection{Multiview Pose Estimation}
\label{sec:gencam_eval}

\parnobf{Experimental Details.}
We evaluate on two scene-scale datasets: RE10K~\cite{zhou2024stereomagnif}, which contains camera trajectories from various video clips, and ETH3D~\cite{schops2017multi}, which contains multi-view indoor and outdoor scenes with high-precision ground-truth geometry and camera poses. Following prior work~\cite{wang2025pi3, wang2025vggt, wang2023posediffusion}, we randomly sample and infer on 10 images per scene and compute metrics on all pairs. 

\parnobf{Metrics.} Following prior work~\cite{wang2025pi3, wang2025vggt, wang2023posediffusion}, we report the $30^\circ$ threshold for relative rotation accuracy (RA$_{30}$), relative translation accuracy (TA$_{30}$), and AUC (AUC$_{30}$).

\parnobf{Evaluation.} Results are reported in Table~\ref{tab:relpose-angular}. As shown, despite fine-tuning on image pairs and a rotation loss alone, rotation and translation accuracies are mostly preserved across all models, indicating that the aligned variants continue to perform well in multi-view settings. While WM and \picubed{} show slight decreases in AUC after fine-tuning, they remain highly competitive—likely nearing their performance ceiling in multi-view pose estimation, where non-overlapping images are rare. In contrast, VGGT exhibits substantial gains across all metrics. We hypothesize that, as the weakest initial model, VGGT has the greatest capacity for improvement, and hence our lightweight alignment scheme most effectively enriches its internal geometric representation. The dense reconstruction results presented next further support this observation.

\begin{table}[t]
    \centering
    \caption{
        \textbf{Dense Reconstruction} on UnSceneRecon and ETH3D~\cite{schops2017multi} datasets. The accuracy (ACC) and completion (CMP) are reported in meters. Non-negligible differences ($>5\%$ relative) between the base and fine-tuned models are in \textbf{bold}.
    }
    \vspace{-0.75em}
    \tablestyle{3pt}{1.0}
    \resizebox{\columnwidth}!{
    \begin{tabular}{l cccc cccc}
        \toprule
        {\multirow{4}{*}{\textbf{Method}}} &
        \multicolumn{4}{c}{\textbf{UnSceneRecon}} &
        \multicolumn{4}{c}{\textbf{ETH3D}} \\
        \cmidrule(r){2-5} \cmidrule(r){6-9}
        &
        \multicolumn{2}{c}{ACC $\downarrow$}  &
        \multicolumn{2}{c}{CMP $\downarrow$} &
        \multicolumn{2}{c}{ACC $\downarrow$}  &
        \multicolumn{2}{c}{CMP $\downarrow$} \\
        \cmidrule(r){2-3} \cmidrule(r){4-5} \cmidrule(r){6-7} \cmidrule(r){8-9}
        &
        Mean & Med. &
        Mean & Med. &
        Mean & Med. &
        Mean & Med. \\
        \midrule

VGGT & 1.441 & 1.049 & 1.403 & 0.729 & 0.284 & 0.194 & 0.342 & 0.204 \\
\vggtft & \textbf{1.291} & \textbf{0.908} & \textbf{1.155} & \textbf{0.650} & \textbf{0.233} & \textbf{0.144} & \textbf{0.259} & \textbf{0.144} \\
        
        \midrule
        
WM & 0.933 & \textbf{0.612} & 0.702 & 0.387 & 0.285 & 0.201 & 0.301 & 0.169 \\
\worldmirrorft & 0.972 & 0.660 & 0.735 & 0.368 & \textbf{0.235} & \textbf{0.167} & \textbf{0.217} & \textbf{0.126} \\

        \midrule

\picubed & \textbf{0.716} & \textbf{0.466} & \textbf{0.635} & \textbf{0.377} & \textbf{0.187} & \textbf{0.125} & 0.210 & \textbf{0.129} \\
\picubedft & 0.791 & 0.517 & 0.689 & 0.403 & 0.202 & 0.137 & 0.219 & 0.137 \\

        \bottomrule
    \end{tabular}
    }
    \label{tab:mv_recon_eth3d_usr}
\end{table}

\subsection{Dense Reconstruction}
\label{sec:recon_eval}

\parnobf{Experimental Details.}
We evaluate on UnSceneRecon and ETH3D. For ETH3D, we follow \picubed and sample every 5th image. For UnSceneRecon, we sample 10 images per scene using a graph-based greedy algorithm to ensure sufficient visual overlap. %
Following prior work~\cite{wang2025pi3, wang2025cut3r}, we aggregate point head predictions to assemble a point cloud per scene, and align the predicted points to the ground truth using the Umeyama algorithm, followed by iterative closest point.

\parnobf{Metrics.}
 Standard accuracy (ACC) and completion (CMP) metrics are reported following prior work~\cite{wang2025pi3, wang2023dust3r, wang2025vggt}. %

\parnobf{Evaluation.} We report dense reconstruction results in Table~\ref{tab:mv_recon_eth3d_usr}. As shown in the table, our fine-tuned models preserve, and in many cases improve, the reconstruction performance of 3DFMs, despite using only a rotation loss and receiving no supervision on dense outputs. %
Fine-tuning VGGT yields the largest gains across both datasets, while WM, which starts from a stronger baseline, shows substantial improvements on ETH3D and only minor variations on UnSceneRecon. In contrast, \picubed{} exhibits some degradation, which we attribute to architectural differences: \picubed lacks a dedicated per-frame camera token like VGGT and WM, forcing camera information to be distributed throughout the image tokens. We hypothesize that a dedicated camera token helps preserve the model's internal langauge when fine-tuning on extreme rotations. Results over DTU~\cite{jensen2014large} %
and 7Scenes~\cite{shotton2013scene} are provided in the supplementary material. These further validate that our alignment scheme enables preserving reconstruction performance.

\subsection{Ablation Study}
\label{sec:ablation}
We perform a series of ablations %
to assess our alignment scheme over two key questions: (1) \textit{Which components should be fine-tuned?} and (2) \textit{How should the backbone be fine-tuned?} 
For all ablations, we report changes in extreme rotation error ($\Delta$\textbf{ROT}) on \textit{UnScenePairs} (using MRE) and changes in reconstruction accuracy ($\Delta$\textbf{REC}) on \textit{UnSceneRecon} (by averaging median ACC and CMP) to evaluate both extreme-view performance and dense reconstruction quality. %
We report results for VGGT and WorldMirror (WM); additional metrics and $\pi^3$ results (over a subset of ablations, as this model lacks depth predictions) are provided in the supplementary material. Overall, we observe that $\pi^3$ follows the same trends observed in WM. %

\begin{table}[t]
    \centering
    \caption{
        Ablation study evaluating rotation ($\triangle$\textbf{ROT}) and reconstruction
        changes relative to pretrained models.
        We report reconstruction changes for both point-head predictions
        ($\triangle$\textbf{REC}\textsubscript{PH}) and depth-fused predictions
        ($\triangle$\textbf{REC}\textsubscript{Fused}).
        Part 1 studies which components to fine-tune.
        The camera head is denoted as $\mathcal{D}_c$ and the backbone as AA. 
        Part 2 ablates update strategies for the AA backbone using
        layer-only (LO), bias-only (BO), or both (LO+BO) updates.
        Significant improvements ($>10\%$ error drop) are shown in
        \textcolor{ForestGreen}{green}, and degradations in \textcolor{BrickRed}{red}.
    }
    \vspace{-0.75em}
    \label{tab:ablation_welp_megause_two_column_params}
    \scriptsize
    \setlength{\tabcolsep}{3pt}
    \renewcommand{\arraystretch}{1.1}
    \resizebox{\columnwidth}{!}{
    \begin{tabular}{l c c c c c c c}
        \toprule
        \textbf{Model} &
        \textbf{Comp.} &
        \textbf{LO} &
        \textbf{BO} &
        $\triangle$\textbf{ROT} &
        $\triangle$\textbf{REC}\textsubscript{PH} &
        $\triangle$\textbf{REC}\textsubscript{Fused} &
        \textbf{\#Params} \\
        \midrule

        \multicolumn{8}{l}{\textit{\textbf{Part 1: Component Ablation }}} \\

        \multirow{3}{*}{VGGT} & $\mathcal{D}_c$            & × & × &
            6.8\% &
            0.0\% &
            0.0\% &
            216.2M \\

         & AA+$\mathcal{D}_c$         & × & × &
            \textcolor{ForestGreen}{-74.3\%} &
            (N/A) &
            \textcolor{BrickRed}{+90.3\%} &
            820.9M \\

         & AA                         & × & × &
            \textcolor{ForestGreen}{-69.8\%} &
            \textcolor{ForestGreen}{-33.7\%} &
            \textcolor{ForestGreen}{-9.2\%} &
            604.7M \\
            \cmidrule{2-8}

        \multirow{3}{*}{WM}   & $\mathcal{D}_c$            & × & × &
            -1.4\% &
            0.0\% &
            0.0\% &
            216.2M \\

           & AA+$\mathcal{D}_c$         & × & × &
            \textcolor{ForestGreen}{-47.5\%} &
            (N/A) &
            \textcolor{BrickRed}{+81.5\%} &
            820.9M \\

           & AA                         & × & × &
            \textcolor{ForestGreen}{-41.6\%} &
            \textcolor{BrickRed}{+14.0\%} &
            \textcolor{BrickRed}{+13.6\%} &
            604.7M \\

        \midrule

        \multicolumn{8}{l}{\textit{\textbf{Part 2: AA-Only Ablation}}} \\

        \multirow{4}{*}{VGGT}& AA                         & × & × &
            \textcolor{ForestGreen}{-69.8\%} &
            \textcolor{ForestGreen}{-33.7\%} &
            \textcolor{ForestGreen}{-9.2\%} &
            604.7M\\
         & AA (LO)                    & \checkmark & × &
            \textcolor{ForestGreen}{-66.7\%} &
            \textcolor{ForestGreen}{-33.3\%} &
            (N/A) &
            100.8M \\

         & AA (BO)                    & × & \checkmark &
            \textcolor{ForestGreen}{-69.7\%} &
            \textcolor{ForestGreen}{-16.7\%} &
            (N/A) &
            0.4M \\

         & AA (LO+BO) &
            \checkmark & \textbf{\checkmark} &
            \textcolor{ForestGreen}{-59.8\%} &
            \textcolor{ForestGreen}{-12.4\%} &
            (N/A) &
            0.07M \\\cmidrule{2-8}

        \multirow{4}{*}{WM}& AA                         & × & × &
            \textcolor{ForestGreen}{-41.6\%} &
            \textcolor{BrickRed}{+14.0\%} &
            \textcolor{BrickRed}{+13.6\%} &
            604.7M \\
           & AA (LO)                    & \checkmark & × &
            \textcolor{ForestGreen}{-40.4\%} &
            \textcolor{BrickRed}{+18.3\%} &
            (N/A) &
            100.8M \\

           & AA (BO)                    & × & \checkmark &
            \textcolor{ForestGreen}{-42.3\%} &
            \textcolor{BrickRed}{+12.9\%} &
            (N/A) &
            0.4M \\

         & AA (LO+BO) &
            \checkmark & \checkmark &
            \textcolor{ForestGreen}{-39.0\%} &
            +2.9\% &
            (N/A) &
            0.07M \\

        \bottomrule
    \end{tabular}
    }
\end{table}

\parnobf{Part 1: Which Components to Finetune?}

\parnobf{Backbone unfreezing is essential.}
Our alignment strategy updates only the alternating attention backbone, refining the encoded 3D representation. 
To verify that this component must remain trainable, we freeze the backbone and fine-tune only the camera head ($\mathcal{D}_c$). 
As shown in the first row of each model block in Table~\ref{tab:ablation_welp_megause_two_column_params}, this setting yields degrading prediction for VGGT and very limited improvement for WM on extreme-view rotation estimation comparing to their pretrained weights. 
Meanwhile, the point-head outputs also remain unchanged because the frozen backbone feeds identical features into the dense prediction head. 
These results prove that the 3D representation is encoded entirely in the backbone rather than in the heads, and therefore the backbone must be unfrozen in order to refine it effectively.

\begin{figure}[t]
    \centering
    \includegraphics[width=0.47\textwidth]{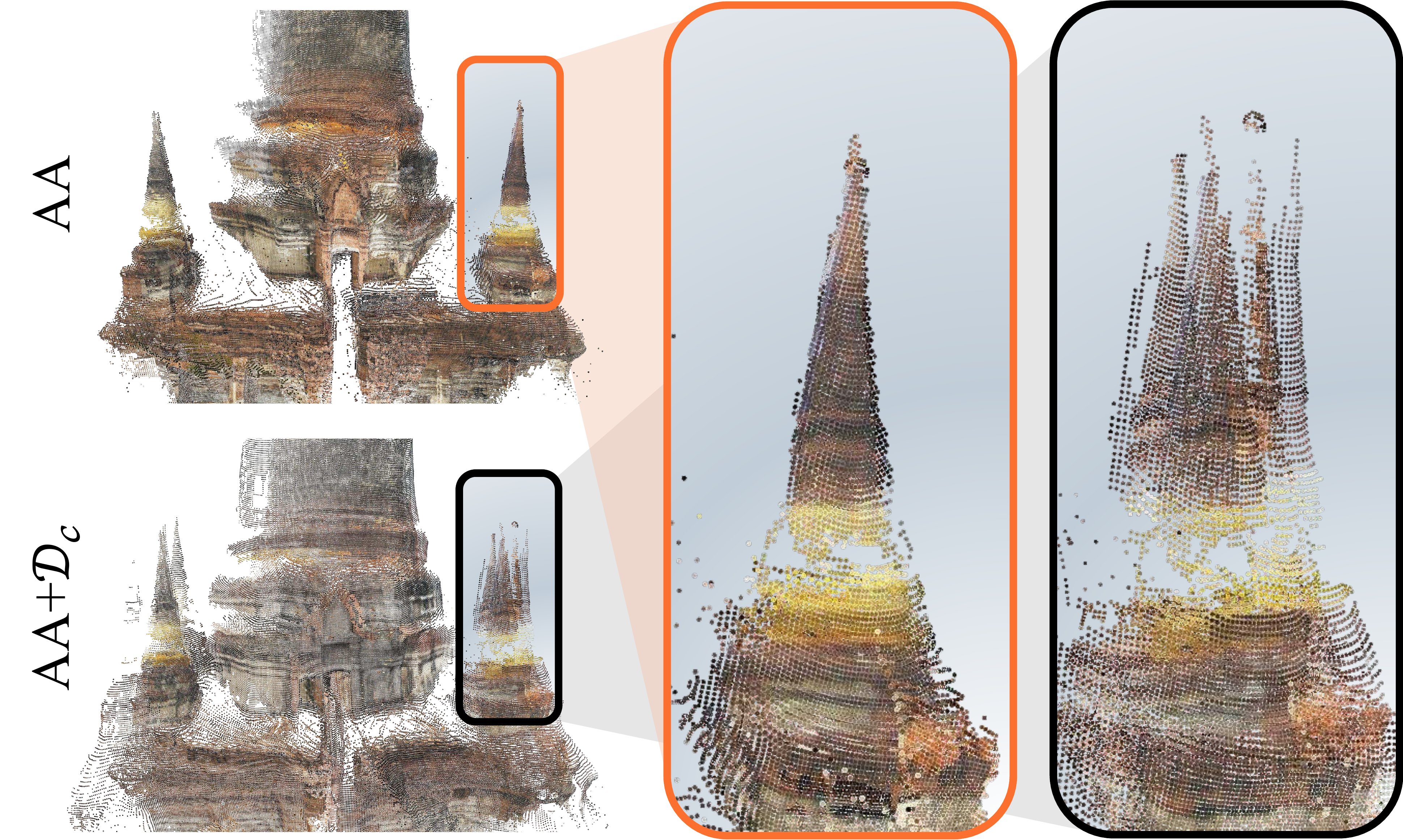}
    \vspace{-8pt}
    \caption{
    Fused VGGT predicted pointmaps from unprojecting depth and applying camera extrinsics on UnSceneRecon's \textit{Wat Yai Chai Mongkhon} scene. The \textbf{\textcolor{myorange}{left zoom-in}} shows better alignment with a frozen camera head (AA), while the \textbf{right zoom-in} shows misalignment with an unfrozen camera head (AA+$\mathcal{D}_c$).
    }
    \label{fig:ablation_cam_head}
\end{figure}

\parnobf{Freezing the camera head preserves multi-task behavior.}
 The camera head should remain frozen to maintain consistency in the model’s shared internal 3D representation between geometry and pose reasoning. To verify this, 
we compare two alignment schemes: (1) unfreezing both the backbone and the camera head, and (2) unfreezing only the backbone. 
As shown in the second and third rows of Table~\ref{tab:ablation_welp_megause_two_column_params}, unfreezing the camera head consistently harms dense reconstruction quality for depth-fused predictions. For instance, freezing the camera head improves VGGT reconstructions by 9.2\% on UnSceneRecon, while unfreezing it causes 90.3\% degradation. %
As shown in Figure~\ref{fig:ablation_cam_head}, modifying the camera head breaks the pretrained depth-pose consistency, highlighting the need to keep it frozen during alignment.

\smallskip\noindent\textbf{Part 2: How to Finetune the Backbone?}

\smallskip \noindent \textbf{Selective bias-only tuning achieves the best trade-off.}
The backbone makes up the majority of a 3DFM's parameters, yet only a small subset needs to be adapted to effectively minimize rotation error. We find that the best options are to (1) tune only biases, in (2) select layers. To demonstrate that both are necessary for preserving the pretrained multi-task knowledge, we compare several configurations varying along these two dimensions: (1) \textit{which layers} are updated, full-layer tuning vs. layer-only (LO) tuning, and (2) \textit{which parameters} are updated, weight-and-bias tuning vs. bias-only (BO) tuning. 
 The results of these four configurations are summarized in the lower half of Table~\ref{tab:ablation_welp_megause_two_column_params}.

Results reveal model-dependent behavior across update strategies. 
For WM, which already exhibit strong pretrained performance in earlier evaluations, 
tuning all weights across layers offers only marginal rotation gains  but noticeably degrades reconstruction quality, indicating overfitting to the rotation objective. 
In contrast, updating only the bias terms of selected layers (LO+BO) achieves a comparable $39.0\%$ reduction in rotation error while keeping reconstruction quality nearly unchanged ($+2.9\%$), using just $0.07$M parameters. 
Well-trained models like WM therefore benefit most from minimal adaptation, whereas models like VGGT with greater capacity for improvement can exploit larger updates to refine its internal 3D representation. 
Overall, these results indicate that fine-tuning only the bias terms of selected backbone layers provides the best balance between geometric refinement and preservation of pretrained multi-task behavior.

\section{Conclusion}

In this work, we study the emergent capabilities of 3D foundation models to understand extreme-view geometry, revealing that they encode a latent 3D language within their alternating attention backbone that can operate beyond visual overlap. We introduce a lightweight alignment scheme that enhances this internal 3D language through selective bias tuning, which modifies only around 80k parameters, four orders of magnitude smaller than the full model. These findings highlight the untapped potential of 3DFMs: even minimal adaptation unlocks strong 3D reasoning across extreme viewpoints, revealing low-parameter alignment as a promising direction for scalable, real-world 3D perception.

{
    \small
    \bibliographystyle{ieeenat_fullname}
    \bibliography{main}
}

\clearpage
\appendix

\twocolumn[
    \begin{center}
        {\LARGE \textbf{Supplementary Material}}\\
    \end{center}
    \vspace{2em}
]

\medskip
\medskip

\label{sec:intro-supp}
We refer readers to the interactive visualizations at the accompanying \texttt{viewer.html} %
that show randomly-selected results for all three 3D foundation models (pre-trained and fine-tuned) on the \textit{relative rotation estimation} and the \textit{dense reconstruction} test sets. In this document, we provide  details regarding our proposed benchmark (Section \ref{sec:dataset-supp}), additional implementation details (Section \ref{sec:details-supp}) and 
provide additional experiments and results (Section \ref{sec:results-supp}).

\section{The MegaUnScene Benchmark}
\label{sec:dataset-supp}

We first provide additional details relating to the curation of MegaUnScene (Section~\ref{sec:supp-initial-curation}). We then provide information on how we construct our three test sets: UnScenePairs (Section~\ref{sec:supp-unscenepairs}), UnScenePairs-t (Section~\ref{sec:supp-unscenepairst}), and UnSceneRecon (Section~\ref{sec:supp-unscenerecon}).

\subsection{Initial Curation}
\label{sec:supp-initial-curation}

We provide additional details on the benchmark construction pipeline for MegaUnScene. First, we describe our construction pipeline, organized in four subsections: identifying scenes, sparse reconstruction, obtaining depth maps, and ensuring unseen scenes. Finally, we summarize MegaUnScene's overall scene statistics.

\parnobf{Scene identification.} We first identify candidate scenes, each corresponding to an image collection, that we want to reconstruct. As mentioned in the main paper, we follow the MegaScenes~\cite{tung2024megascenes} dataset curation pipeline to find scenes and their corresponding image collection from Wikimedia Commons and its sister site, Wikidata. In order to avoid scene overlap with the MegaScenes dataset, we query Wikidata with different high-level classes than in those used in MegaScenes. We filter out all Wikimedia Commons categories whose names intersect with MegaScenes, as category names are unique. This results in approximately 340,000 candidate scenes. For each scene, we follow MegaScenes and download images from all Wikimedia Commons subcategories with a max depth of four.

\parnobf{Sparse Reconstruction}. We then reconstruct each candidate scene with at least 50 images
using Doppelgangers++ (DGPP)~\cite{xiangli2025doppelgangersimprovedvisualdisambiguation} integrated with MASt3R-SfM~\cite{duisterhof2025mastrsfm} as mentioned in the main paper.
In this pipeline, MASt3R~\cite{leyor2024mast3r} is used for image retrieval and matching, followed by match pruning using the DGPP classifier with a threshold of 0.8. We use COLMAP~\cite{schonberger2016structure} for incremental SfM, manhattan-world alignment, and image undistortion. As Internet photos are noisy, not all images in the scene's image collection are registered to a reconstruction; thus we filter again for reconstructions that contain at least 50 images.

\parnobf{Obtaining Depth Maps.} We obtain semi-dense depth maps with COLMAP's stereo fusion~\cite{schoenberger2016mvs}. As described in MegaDepth~\cite{li2018megadepth}, multi-view stereo depth maps typically contain artifacts from reconstruction ambiguities, especially on unconstrained internet photos. We clean these depth maps by following the same depth refinement protocol in MegaDepth. The protocol is slightly modified by replacing the PSPNet~\cite{zhao2017pspnet} segmentation model with the more recent SegFormer~\cite{xie2021segformer} in the semantic filtering step. For more details, please refer to Algorithm 1 of MegaDepth's supplementary material.

\parnobf{Ensuring Unseen Scenes.} As a postprocessing step after reconstruction, we check for image conflicts between \evaldataset and MegaScenes's base 9M image set (a superset of MegaScenes' 2M image subset that is reconstructed). This is possible as Wikimedia Commons ensures that each image has a unique filename. We only use \evaldataset reconstructions with no image conflicts for UnSceneRecon, and reconstructions with less than 10\% image conflict for both \wELP{} and \wELPt{}. For UnScenePairs, we only select image pairs where neither image intersects with MegaScenes. For release, we note all conflicting images registered to reconstructions.

\parnobf{Benchmark statistics.} From the dataset curation pipeline, we identify 758 reconstructions across 658 scenes with $\ge50$ images and $<10\%$ image overlap with MegaScenes. Of these, we release 485 reconstructions across 476 scenes to be used for evaluation in our three new test sets: UnScenePairs, UnscenePairs-t, and UnSceneRecon. A breakdown of dataset statistics is provided in Table~\ref{tab:dataset-statistics}.

\subsection{\wELP{} Test Set}
\label{sec:supp-unscenepairs}
Prior work~\cite{bezalel2025extreme} introduced the wELP (``in-the-wild" Extreme Landmark Pairs) test set curated from MegaDepth~\cite{li2018megadepth}. However, all three 3D foundation models that we fine-tuned were pretrained on MegaDepth. To provide evaluation on a distinct data source but in the same camera-centric distribution, we  follow the same pipeline to filter image pairs on \evaldataset.

As introduced in our paper, the pipeline identifies image pairs with negligible translation and predominant rotation using mutual $K$-nearest neighbor graphs ($K=5$) constructed from the distances between camera translation. Mutual neighbors ensure that only pairs that are consistently close in translation space are preserved. Each surviving image pair is then assigned an overlap level (Large, Small, and None) with the following algorithm:

Given the relative rotation matrix $\mathbf{R}$ between two cameras and their respective FoVs, the overlap category $o$ is determined by:

\begin{equation}
o = \begin{cases}
    \textit{Large} & |\gamma| < \frac{\text{fov}_x^1+\text{fov}_x^2}{4} \land |\beta| < \frac{\text{fov}_y^1+\text{fov}_y^2}{4} \\
    \textit{None} & |\gamma| > \frac{\text{fov}_x^1+\text{fov}_x^2}{2} \land |\beta| > \frac{\text{fov}_y^1+\text{fov}_y^2}{2} \\
    \textit{Small} & \text{otherwise}
\end{cases}
\end{equation}

\noindent where $\gamma$ and $\beta$ denote the relative yaw and pitch angles extracted from $\mathbf{R}$ using Euler decomposition. 

After the pipeline, we manually review all selected pairs and removed those affected by motion blur, occlusions, or insufficient geometric structure. The resulting \wELP{} statistics is shown in Table~\ref{tab:dataset-statistics}.

\begin{table}[t]
\caption{\textbf{MegaUnScene Statistics.} Statistics for our benchmark and three MegaUnScene test sets: \wELP{}, \wELPt{}, and UnSceneRecon. For \wELP{}, and \wELPt{}, we report the number of image pairs extracted for each overlap level. For UnSceneRecon, we report the number of reconstructions. At the bottom, we summarize the total number of unique MegaUnScene scenes and reconstructions across the three test sets.}
\vspace{-8pt}
\setlength{\tabcolsep}{3.8pt}
\def\arraystretch{0.95}
\centering
\resizebox{0.48\textwidth}{!}{%
\begin{tabular}{lrrrrrr}
\toprule 
& & & \multicolumn{4}{c}{\#Pairs} \\ 
\cline{4-7} 
Subset & \# Scenes & K & Large & Small & None & Total  \\ 
\midrule
\wELP{} & 458 & 5 & 1,878 & 1,227 & 778 & 3,883 \\
\wELPt{} & 387 & 50 & 1,146 & 523 & 763 & 2,432 \\

\midrule
& & & \\ 
Subset & \# Scenes & \# Recons & \multicolumn{4}{c}{Notes} \\
\midrule
UnSceneRecon & 96 & 100 & \multicolumn{4}{c}{Human-annotated metric scale} \\

\midrule
& & & \\ 
Overall & \# Scenes & \# Recons & \multicolumn{4}{c}{Notes} \\
\midrule
MegaUnScene & 476 & 485 & \multicolumn{4}{c}{\text{Counts after de-duplication}} \\
\bottomrule
\end{tabular}
}

\label{tab:dataset-statistics}
\end{table}

\subsection{\wELPt{} Test Set}
\label{sec:supp-unscenepairst}

As discussed in our paper, we also construct \textit{\wELPt{}} from \evaldataset with the same pipeline but use $K=50$ mutual nearest neighbors to evaluate performance on pairs with larger camera translations. We then implement correspondence-based verification using Doppelgangers++~\cite{xiangli2025doppelgangersimprovedvisualdisambiguation} + MASt3R-SfM~\cite{duisterhof2025mastrsfm}'s reconstructed geometry and checking the verified inlier matches.

Additionally, with larger camera translations included as a consequence of setting $K=50$, the same scene structure may appear at vastly different scales depending on camera-to-scene distance, where a telephoto lens from afar and a wide-angle lens nearby could capture the same 3D structure at incompatible image resolutions. To ensure scale consistency, we extend the basic FoV threshold with three criteria: (1) \textit{both} horizontal and vertical FoV differences below 15° independently, (2) focal length ratio $\max(f_x^1, f_x^2) / \min(f_x^1, f_x^2) < 2.5$ to catch zoom differences, and (3) image resolution ratio $\max(w_1 h_1, w_2 h_2) / \min(w_1 h_1, w_2 h_2) < 3.0$ to prevent sensor size discrepancies from obscuring focal length mismatches.

Finally, similar to \wELP{} filtering, we also manually inspected \wELPt{} and removed noisy image pairs. Statistics are shown in Table~\ref{tab:dataset-statistics}.

\subsection{UnSceneRecon Test Set}
\begin{figure*}[t]
    \centering
    \includegraphics[width=\linewidth]{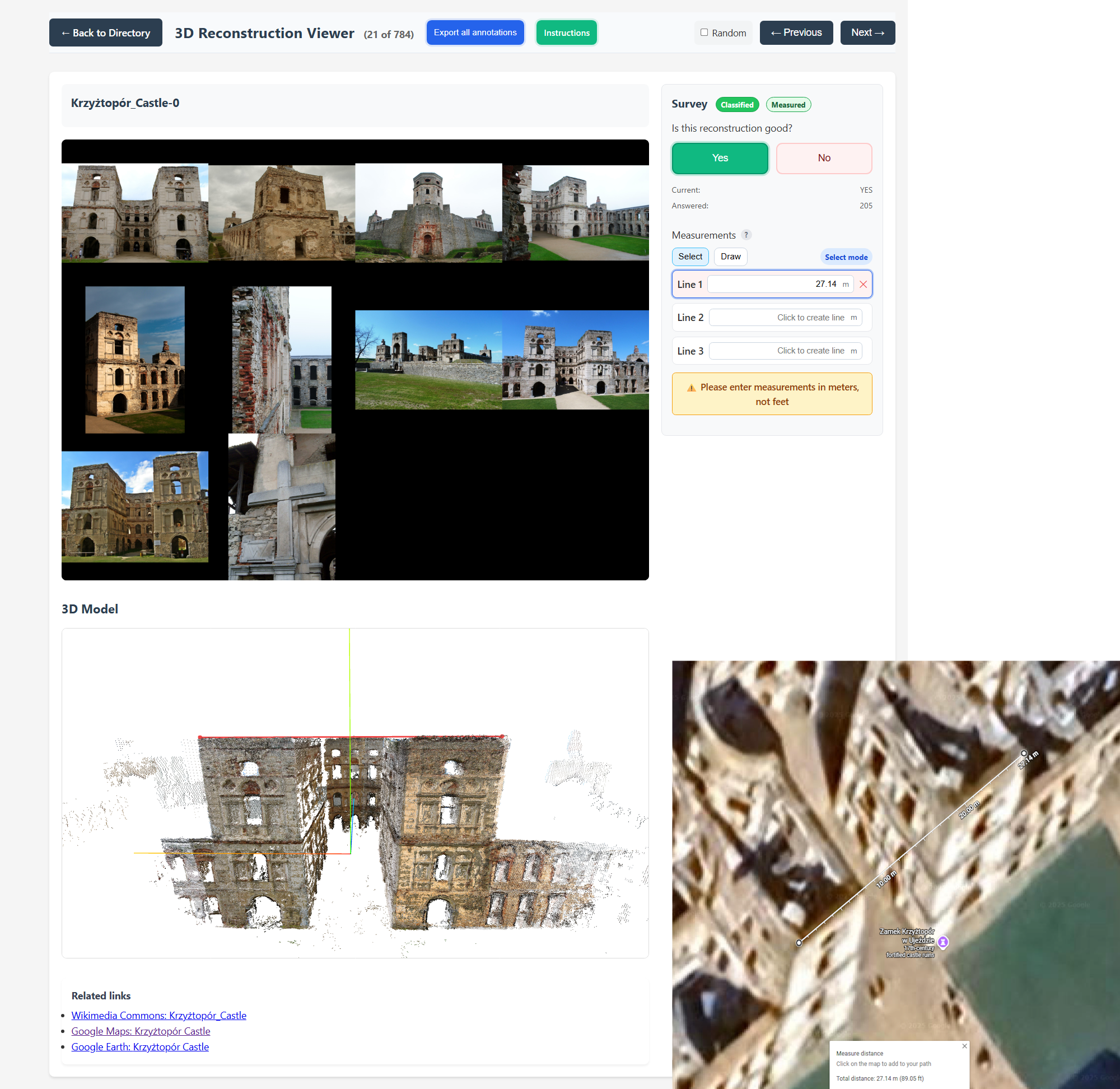}
    
    \caption{\textbf{UnSceneRecon Reconstruction Annotator}. We show the annotator webpage on MegaUnScene's Krzyżtopór Castle scene (left) and its corresponding Google Maps satellite view (bottom right). At the top-left of the viewer, we depict a randomly sampled set of 10 images. On the bottom-left, we show the corresponding 3D model from unprojecting the depths of these 10 images in the global coordinate frame. On the top right of the page, annotators label whether the reconstruction is good or not, and also make a metric estimate of the reconstruction. The metric estimate is done by drawing a line on the reconstruction (shown at the bottom-left, at the top of the building's 3D model in red), measuring the corresponding distance in Google Maps (shown on the bottom right), and pasting the measurement in the corresponding field on the top-right of the viewer. In this example, the annotator labeled ``Yes" that the reconstruction is good, and that the annotated red line is 27.14 meters. We provide links to the Wikimedia Commons page, as well as a Google Maps page that searches the scene name, to help annotators identify the correct location on Google Maps.
    }
    \label{fig:unscenerecon_annotator}
    \vspace{1.5cm}
\end{figure*}

\label{sec:supp-unscenerecon}

We construct a user interface for human annotators to annotate each reconstruction, as depicted in Figure~\ref{fig:unscenerecon_annotator}. Annotators are first instructed to visually assess reconstruction quality to see determine if the reconstruction is realistic based on the images; they label the reconstruction ``good" or ``bad" accordingly. If a reconstruction is good, they  are instructed to draw a line on the reconstruction, find the corresponding points on Google Maps in satellite view, and annotate the metric scale (as shown on Figure~\ref{fig:unscenerecon_annotator}). In practice, we instruct annotators to only label one line to estimate the metric scale. From this process, we label 100 reconstructions across 96 scenes with metric scale annotations, as shown in Table~\ref{tab:dataset-statistics}.

\section{Implementation Details}
\label{sec:details-supp}
We first describe how we select backbone layers for fine-tuning (Section \ref{sec:supp-layer-selection}), then outline the construction of the training set (Section \ref{sec:supp-train-set}), followed by our training configuration (Section \ref{sec:supp-train-details}), and finally provide the full evaluation protocols used across all tasks (Section \ref{sec:supp-eval-details}).

\subsection{Backbone Layer Selection}
\label{sec:supp-layer-selection}
\begin{figure*}[t]
    \centering

    \begin{minipage}[c]{0.03\textwidth}
        \centering
        \rotatebox{90}{Similarity}
    \end{minipage}
    \hspace{-12pt}
    \begin{minipage}[c]{0.95\textwidth}
        \centering
        \begin{minipage}{0.32\linewidth}
            \centering
            \hspace*{10pt}VGGT\\[0mm]
            \includegraphics[height=3cm]{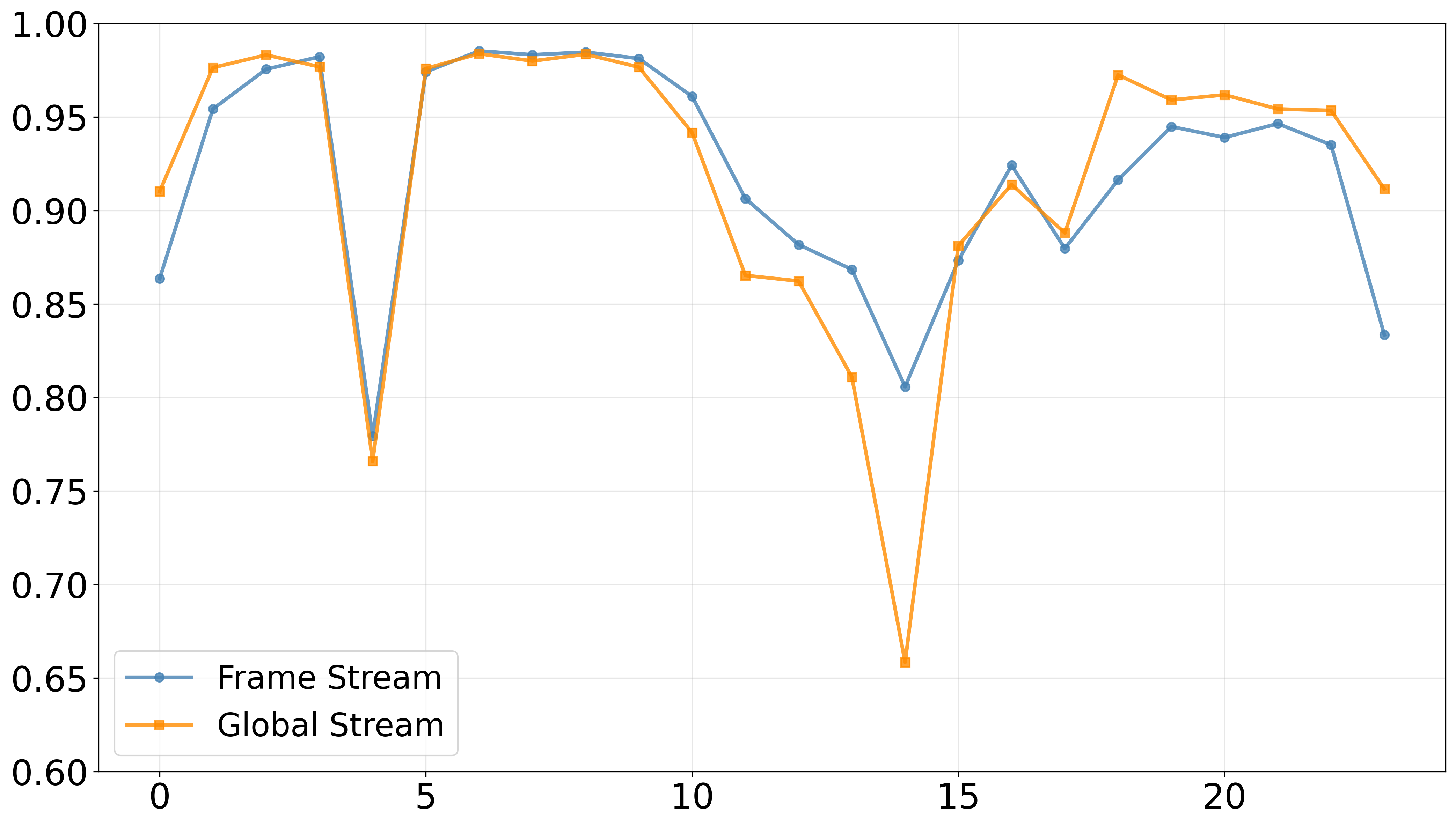}
        \end{minipage}
        \begin{minipage}{0.32\linewidth}
            \centering
            \hspace*{10pt}WM\\[0mm]
            \includegraphics[height=3cm]{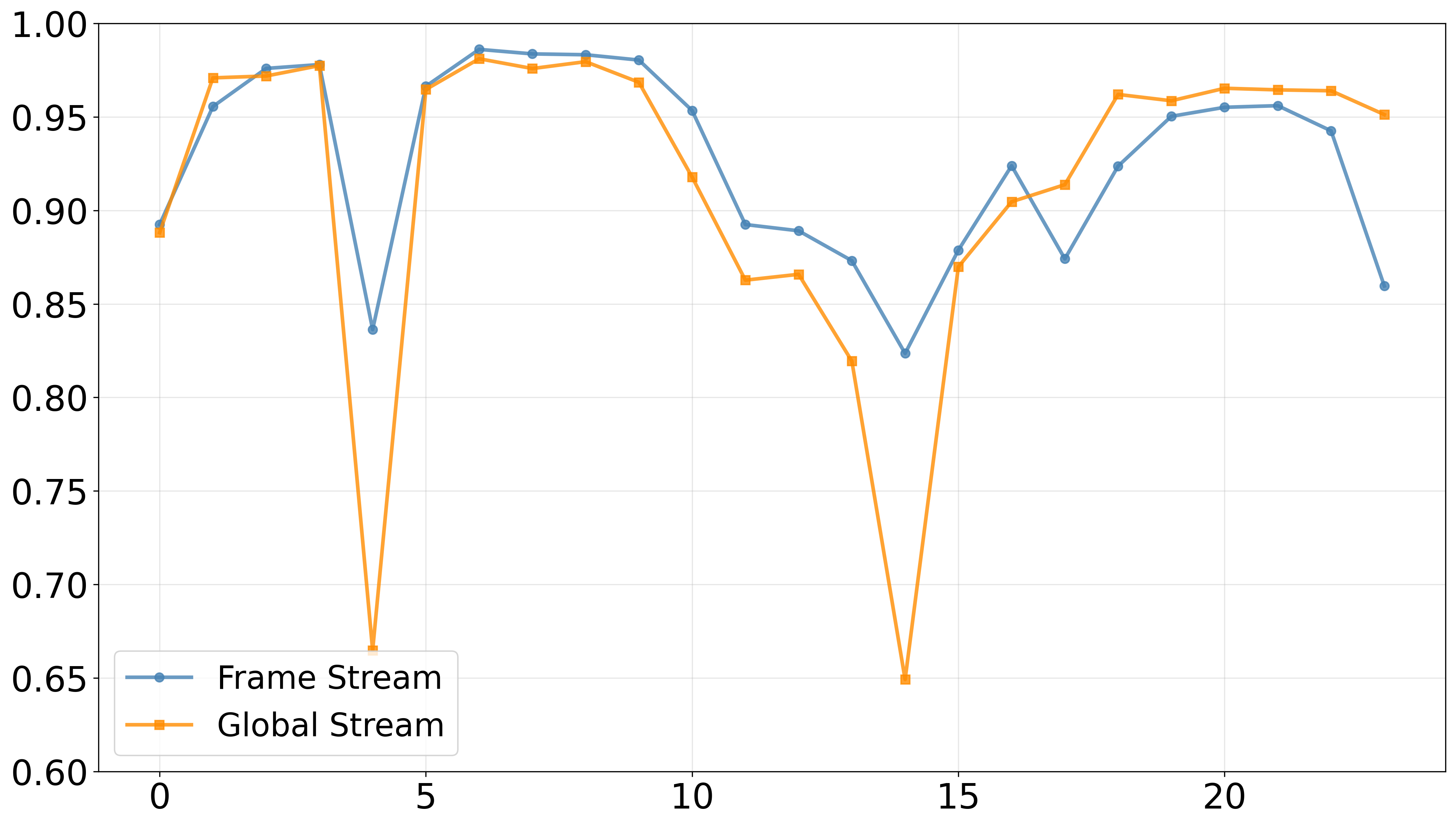}
        \end{minipage}
        \begin{minipage}{0.32\linewidth}
            \centering
            \hspace*{10pt}$\pi^3$\\[0mm]
            \includegraphics[height=3cm]{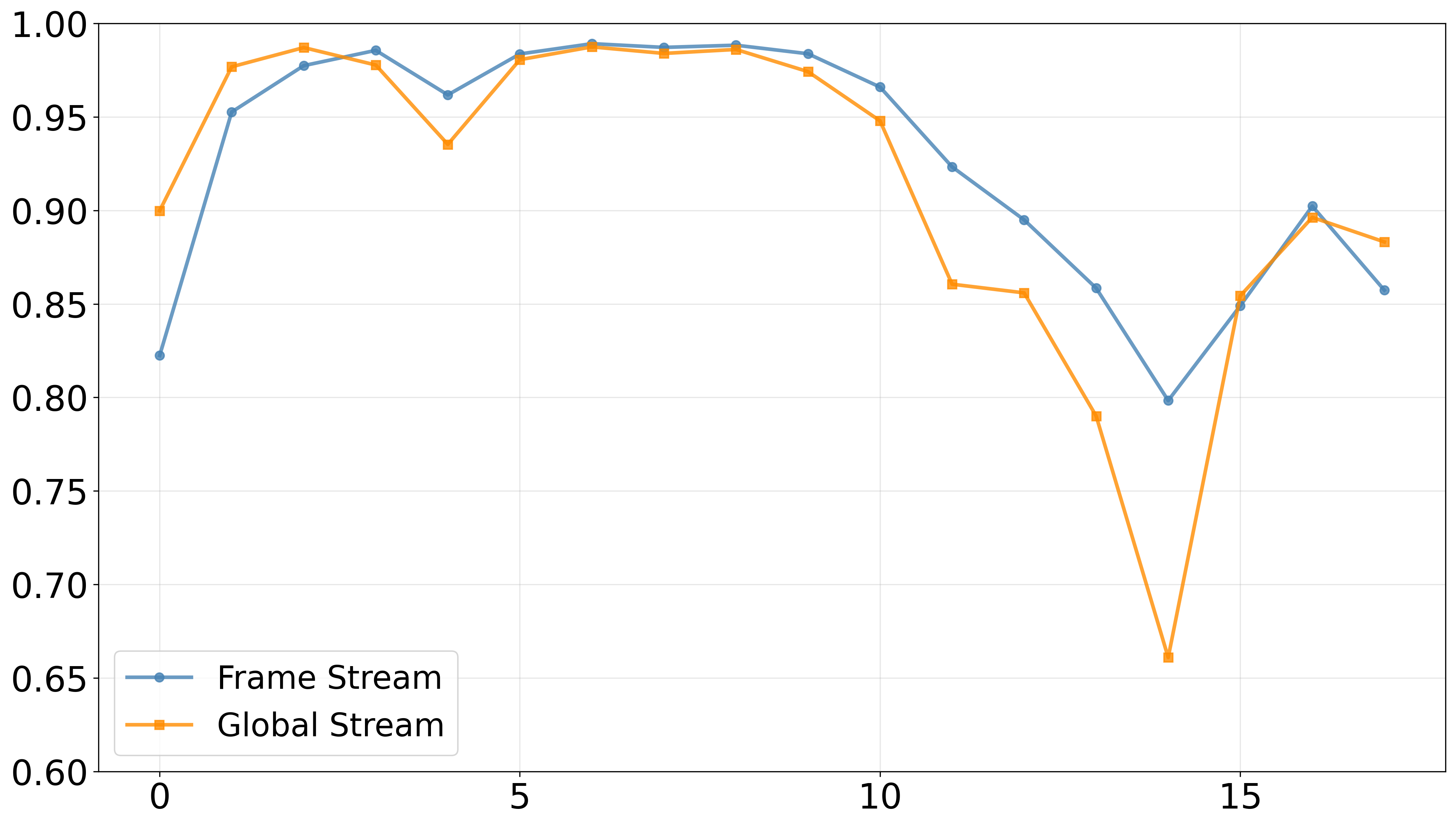}
        \end{minipage}
        \vspace{-4pt}
        Layer Index
    \end{minipage}

    \caption{
        \textbf{Layer Analysis.} Cosine similarity curves of the input–output representations for pretrained VGGT, WorldMirror (WM), and \picubed. For VGGT and WM, the curves span 24 layers, where each layer corresponds to a frame-global block pair. For \picubed, the curve spans 18 layers. Our fine-tuning focuses on layers with pronounced similarity drops; see Section \ref{sec:supp-layer-selection} for additional details. %
    }
    \label{fig:token}
\end{figure*}

To quantify the degree of representational change between neighboring backbone layers, 
we follow the layer similarity pipeline conducted by prior work~\cite{he2024matterstransformersattentionneeded} and run forward passes on ten image pairs, measuring the similarity between the input and output token representations of each layer. The input and output tokens $\mathbf{T}_l^{\text{in}}$ and $\mathbf{T}_l^{\text{out}}$ correspond to either the frame tokens $\mathbf{T}_{\text{frame}}^{(i)}$ or the global tokens $\mathbf{T}_{\text{global}}$, as defined in the method section. For a layer $l$, we compute the cosine similarity:
\begin{equation}
\text{sim}_l = \frac{\mathbf{T}_l^{\text{in}} \cdot \mathbf{T}_l^{\text{out}}}{\lVert \mathbf{T}_l^{\text{in}} \rVert \cdot \lVert \mathbf{T}_l^{\text{out}} \rVert}.
\end{equation}

We run the pipeline on the frame and global attention blocks of VGGT, WorldMirror (WM), and \picubed\ separately, and the resulting similarity curves are shown in Figure~\ref{fig:token}. As can be observed in the figure, the curves of WM and VGGT exhibit similar similarity drops, which is expected given that WM inherits both the architecture and weight initialization of VGGT. We also find that the similarity drop regions—where the curve shows a pronounced decline—coincide with the intermediate layers commonly used for dense predictions, namely layers 4, 11, 17, and 23. We therefore adopt this fixed set for these models. %

For \picubed\ which doesn't include such skip connections, we select layers by detecting
local minima in the similarity curve (using peak detection on the inverted signal) and expanding around each minimum to include adjacent layers with low similarity scores. This ensures that both the most transformative layers and their contextually relevant neighbors are captured. The selection criterion is defined as: \begin{equation} \mathcal{L} = \bigcup_{i \in \mathcal{M}} \left\{ i \pm k : s_{i \pm k} \leq \bar{s} - \frac{\sigma_s}{2}, \, k \in [1, \delta] \right\}, \end{equation} where $\mathcal{M}$ is the set of detected local minima, $\bar{s}$ and $\sigma_s$ are the mean and standard deviation of similarity scores, and $\delta$ controls the neighborhood expansion radius. In practice, we use $\delta = 2$. This yields a selected subset which includes frame layers 4, 12–16 and global layers 13–15.

\subsection{MegaScenes Train Set}
\label{sec:supp-train-set}
For the train set, we use the same pipeline described in Section~\ref{sec:supp-unscenepairs} with $K=50$ to filter image pairs from scene-level COLMAP reconstructions in MegaScenes~\cite{tung2024megascenes} and employ a balanced subsampling strategy to ensure uniform pair selection across overlap categories. We first cap each scene at a maximum of 40 pairs to prevent scene-level bias, then subsample to achieve exact balance across the three overlap categories. The final train set contains 64,584 image pairs (21,528 for each overlap category) across 3,284 scenes.

\subsection{Training Details}
\label{sec:supp-train-details}
We use the same training configuration for all three models. Training is performed with the AdamW optimizer using a learning rate of $5 \times 10^{-5}$ and a weight decay of $1 \times 10^{-4}$. We train on four NVIDIA RTX A6000 GPUs in a distributed setting with a per-GPU batch size of $1$.
For each selected layer, we update only the bias parameters in the attention and MLP modules---specifically, the biases of the query--key--value projection ($\text{attn.qkv.bias}$), the attention output projection ($\text{attn.proj.bias}$), and the two MLP fully connected layers ($\text{mlp.fc1.bias}$ and $\text{mlp.fc2.bias}$).
To stabilize training, we apply gradient clipping with a maximum norm of $1.0$ to mitigate gradient explosion.

Additionally, for the ablation checkpoints that unfreezes both weights and biases of a backbone layer or the camera head, we apply LoRA~\cite{hu2021loralowrankadaptationlarge} with rank = 24, alpha = 48, and dropout = 0.1 for parameter efficient fine-tuning.

\subsection{Evaluation Protocols}
\label{sec:supp-eval-details}
In this subsection, we provide additional details on our evaluation protocols and settings.

\parnobf{General Evaluation Settings.}
For all non-ablation evaluations, the base models we use for VGGT, WorldMirror, and \picubed are the publicly available checkpoints from HuggingFaceHub with the following names:

\begin{itemize}
    \item \texttt{facebook/VGGT-1B}
    \item \texttt{tencent/HunyuanWorld-Mirror}
    \item \texttt{yyfz233/Pi3}
\end{itemize}

The fine-tuned models are the Layer-Only (LO) and Bias-Only (BO) checkpoints.

\parnobf{Relative Rotation Evaluation Settings.}
We preprocess input images using each architecture's provided functions: WorldMirror and VGGT crop images to width 518; \picubed proportionally scales with a pixel limit, preserving aspect ratio. All three ensure image width and height are multiples of 14 after preprocessing. The predicted quaternions are converted into rotational matrices for evaluation. 

For ExRot~\cite{bezalel2025extreme}, we evaluate on their publicly available model on their GitHub page. For all datasets, the images are downsized such that the longer dimension is 256 pixels, then center zero-padded to be size 256x256.

\paragraph{Multiview Pose Estimation, Monocular Depth, and Dense Reconstruction Settings.}
We also preprocess input images with each architecture's provided function. For multview pose estimation and monocular depth, we follow \picubed's~\cite{wang2025pi3} protocol and downsize images to a target size of 512 pixels, with dimensions adjusted to be divisible by 14 through rounding. For dense reconstruction, the target size is 518 pixels. As UnSceneRecon is the only dataset with variable aspect ratio, we downsize the longest edge to 518 and center zero-pad to be 518x518 pixels for the dense reconstruction evaluation.

\parnobf{UnSceneRecon Graph-Based Image Sampling for Dense Reconstruction Evaluation.}
In the main paper, we mention that we subsample images from UnSceneRecon using a graph-based greedy algorithm for dense reconstruction evaluations.
This is because UnSceneRecon scenes typically have widely distributed camera poses that capture different portions of the scene. Random image sampling often leads to poor overlap---\ie, selecting images from disjoint locations on opposite sides of the scene---that are implausible for reconstruction pipelines to reasonably reconstruct. Our graph-based approach ensures connectivity across images while maintaining diversity.

We construct an image connectivity graph for each sparse reconstruction, where nodes represent images and edges connect image pairs with at least 30 shared 3D points and a translation of at least 5 meters. We initialize with a random node in the largest connected component, then greedily sample images based by selecting neighbors that maximize a score combining 80\% connectivity and 20\% diversity. Here, ``connectivity" is the normalized node degree (degree divided by the maximum degree in the graph); ``diversity" is the average distance from a candidate to the nodes of all selected images, normalized by the maximum translation across all edges in the graph.

\section{Experiments and Results}
\label{sec:results-supp}
We begin by evaluating relative camera pose across both overlapping and non-overlapping settings (Section \ref{sec:supp-relrot}), then present additional dense reconstruction experiments on multiple benchmarks (Section \ref{sec:supp-dense-recon}), followed by monocular depth evaluations (Section \ref{sec:monodepth_eval}), and conclude with expanded ablation studies analyzing alternative fine-tuning strategies (Section \ref{sec:supp-extra-ablations}).

\subsection{Relative Camera Pose}
\label{sec:supp-relrot}
\begin{table*}[t]
\caption{Expanded comparison of sELP, UnScenePairs, and UnScenePairs-t benchmarks across VGGT, WorldMirror (WM), \picubed, and their fine-tuned variants. MRE and MTE report the median rotation and translation errors in degrees. RA$_{15}$/RA$_{30}$ and TA$_{15}$/TA$_{30}$ indicate the percentage of predictions whose rotation or translation errors are below 15° or 30°, respectively. }
\setlength{\tabcolsep}{3pt}
\def\arraystretch{1}
\centering
\resizebox{\textwidth}{!}{%
\begin{tabular}{llccc|ccc|cccccc}
\toprule
\multicolumn{2}{c}{} &
\multicolumn{3}{c}{sELP} &
\multicolumn{3}{c}{UnScenePairs} &
\multicolumn{6}{c}{UnScenePairs-t} \\
\cline{3-5} \cline{6-8} \cline{9-14}
Overlap & Method &
MRE & RA$_{15}$ & RA$_{30}$ &
MRE & RA$_{15}$ & RA$_{30}$ &
MRE & RA$_{15}$ & RA$_{30}$ &
MTE & TA$_{15}$ & TA$_{30}$ \\
\midrule

\multirow{6}{*}{Large} &
VGGT &
0.75 & 99.7 & 99.7 &
0.99 & 99.7 & 99.8 &
1.07 & 99.9 & 100.0 & 2.44 & 92.7 & 97.0 \\

& \vggtft &
0.95 & 99.9 & 100.0 &
1.04 & 99.6 & 99.7 &
1.08 & 99.9 & 99.9 & 2.24 & 94.0 & 97.6 \\

& WM &
0.58 & 99.7 & 99.7 &
1.23 & 96.8 & 99.7 &
1.20 & 99.0 & 99.7 & 3.08 & 87.4 & 94.3 \\

& \worldmirrorft &
0.82 & 99.7 & 99.7 &
1.40 & 98.2 & 99.5 &
1.15 & 99.4 & 99.7 & 3.34 & 88.7 & 94.6 \\

& \picubed &
0.68 & 99.8 & 99.8 &
1.24 & 96.8 & 99.2 &
1.09 & 99.2 & 99.8 & 3.39 & 86.7 & 94.1 \\

& \picubedft &
0.93 & 99.9 & 99.9 &
1.23 & 99.4 & 99.8 &
0.99 & 99.9 & 100.0 & 3.05 & 88.7 & 95.9 \\

\midrule

\multirow{6}{*}{Small} &
VGGT &
1.86 & 96.9 & 97.8 &
1.93 & 96.7 & 98.0 &
2.01 & 98.8 & 99.8 & 9.69 & 62.7 & 74.0 \\

& \vggtft &
2.12 & 98.8 & 99.4 &
1.95 & 97.5 & 98.8 &
2.04 & 100.0 & 100.0 & 9.18 & 61.4 & 75.1 \\

& WM &
1.20 & 97.7 & 98.3 &
2.37 & 95.4 & 98.6 &
1.98 & 97.7 & 99.8 & 11.15 & 55.8 & 72.8 \\

& \worldmirrorft &
1.92 & 98.1 & 98.2 &
2.65 & 96.5 & 98.9 &
2.39 & 97.9 & 99.6 & 11.75 & 56.4 & 72.1 \\

& \picubed &
1.50 & 97.3 & 97.7 &
2.31 & 94.4 & 98.5 &
1.77 & 98.1 & 99.8 & 9.68 & 60.6 & 75.3 \\

& \picubedft &
2.29 & 98.4 & 98.7 &
2.43 & 96.7 & 99.1 &
1.99 & 99.0 & 100.0 & 10.91 & 57.5 & 75.5 \\

\midrule

\multirow{6}{*}{None} &
VGGT &
92.92 & 24.2 & 29.1 &
31.64 & 33.8 & 48.8 &
46.65 & 29.1 & 42.1 & 37.28 & 25.3 & 42.2 \\

& \vggtft &
14.21 & 50.9 & 56.5 &
12.71 & 53.6 & 67.9 &
14.48 & 50.6 & 62.1 & 35.79 & 26.5 & 44.0 \\

& WM &
68.96 & 36.3 & 42.5 &
19.25 & 44.1 & 58.9 &
21.52 & 42.6 & 57.4 & 33.83 & 27.6 & 45.2 \\

& \worldmirrorft &
9.74 & 56.9 & 63.5 &
11.75 & 56.2 & 68.1 &
13.13 & 53.3 & 64.5 & 33.42 & 27.8 & 46.7 \\

& \picubed &
45.24 & 43.8 & 48.3 &
17.66 & 46.5 & 59.4 &
21.62 & 43.5 & 56.8 & 33.16 & 29.4 & 45.7 \\

& \picubedft &
11.96 & 53.7 & 60.0 &
12.92 & 54.0 & 69.2 &
13.31 & 53.1 & 65.5 & 32.05 & 31.4 & 47.7 \\

\bottomrule
\end{tabular}
}
\label{tab:all_relpose}
\end{table*}

\parnobf{Evaluations on Large/Small Overlapping Pairs.}
As shown in Table~\ref{tab:all_relpose}, we also evaluate the three fine-tuned models on large and small overlapping image pairs from \cambridge, \wELP{}, and \wELPt{}. All models achieve comparable, and sometimes slightly improved, rotation accuracy. This demonstrates that our fine-tuning procedure does not compromise performance on overlapping image pairs. Since the pretrained models already produce strong rotation estimates when overlap is present, fine-tuning preserves this capability.

\parnobf{Translation Evaluation on \wELPt{}.}
For \wELPt{}, ground-truth relative translations are also available. 
Following prior work~\cite{wang2024posediffusionsolvingposeestimation}, 
we evaluate translation accuracy using the angular error. 
Let $\mathbf{t}_{21}$ and $\mathbf{t}_{21}^{\star}$ denote the predicted 
and ground-truth translation vectors from camera~2 to camera~1. 
The error is defined as
\begin{equation}
\mathrm{err}_T 
= 
\arccos\!\left(
\frac{
\bigl|\mathbf{t}_{21}^{\top}\mathbf{t}_{21}^{\star}\bigr|
}{
\|\mathbf{t}_{21}\|\;\|\mathbf{t}_{21}^{\star}\|
}
\right).
\end{equation}

As shown in Figure~\ref{tab:all_relpose}, we report the Median Translation Error (MTE) and the Translation Accuracy (TA) at thresholds of $15^\circ$ and $30^\circ$, denoted $\mathrm{TA}_{15}$ and  $\mathrm{TA}_{30}$. Since our loss only supervises over the predicted rotation, the translation error and accuracy stay the roughly the same after fine-tuning, with small improvement on the non-overlapping pairs for all three models. As discussed in the paper, this suggests that all fine-tuning framework performs minimal updates to the pretrained weights without overfitting to predicting extreme relative rotation.

\parnobf{Supervising over Full Pose.}
We also experiment with adding a loss term to supervise relative translation in addition to rotation. 
Each of the three models addresses scale ambiguity differently: 
VGGT normalizes all camera translations in the training data, 
WorldMirror provides a normalized camera pose through prior prompting, 
and \picubed{} computes a scale factor by aligning its predicted point map with 
the ground truth. We follow VGGT by normalizing each translation vector using the mean distance from the ground truth sparse 3D points to the point-cloud center, noting that VGGT measures distances to the origin while our dataset does not anchor the first image. Additionally for \picubed{}, we adopt a similar scaling strategy. However, with only two input images, aligning a partial predicted point map to the full ground-truth scene is unstable. We instead apply a simple scaling heuristic. 
For each image pair \(i\), let
\(\mathbf{t}_{\mathrm{pred}}^{i}\) and \(\mathbf{t}_{\mathrm{gt}}^{i}\) denote
the predicted and ground-truth relative translations in world coordinate.
We compute a scale factor
\begin{equation}
    s_i =
    \frac{\bigl\|\mathbf{t}_{\mathrm{gt}}^{i}\bigr\|_2}
         {\bigl\|\mathbf{t}_{\mathrm{pred}}^{i}\bigr\|_2},
\end{equation}
and rescale the prediction as
\begin{equation}
    \tilde{\mathbf{t}}_{\mathrm{pred}}^{\,i}
    = s_i \,\mathbf{t}_{\mathrm{pred}}^{i},
\end{equation}

We add an L1 loss on the translation vectors and keep the same geodesic loss for rotations. For VGGT and WorldMirror, the translation loss anchors the first image as the reference frame and compute the loss on the two absolute predicted translations where the first image's translation should be [0,0,0] in world coordinate. For \picubed{}, we instead compute the loss on the relative translation vectors. Results in Table~\ref{tab:og_loss} show that this additional translation supervision provides essentially no improvement in translational or rotational accuracy, and occasionally will lead to worse performances. This illustrates that our proposed rotation-based objective can better align models to extreme-view geometries, in comparison to the full pose objective used by prior work. As mentioned in the paper, our finding also shows that predicting large translational displacements between two images is intrinsically hard and remains to be an important future line of work.

\subsection{Dense Reconstruction}
\label{sec:supp-dense-recon}
\begin{figure*}[t]
    \centering
    \includegraphics[width=\linewidth]{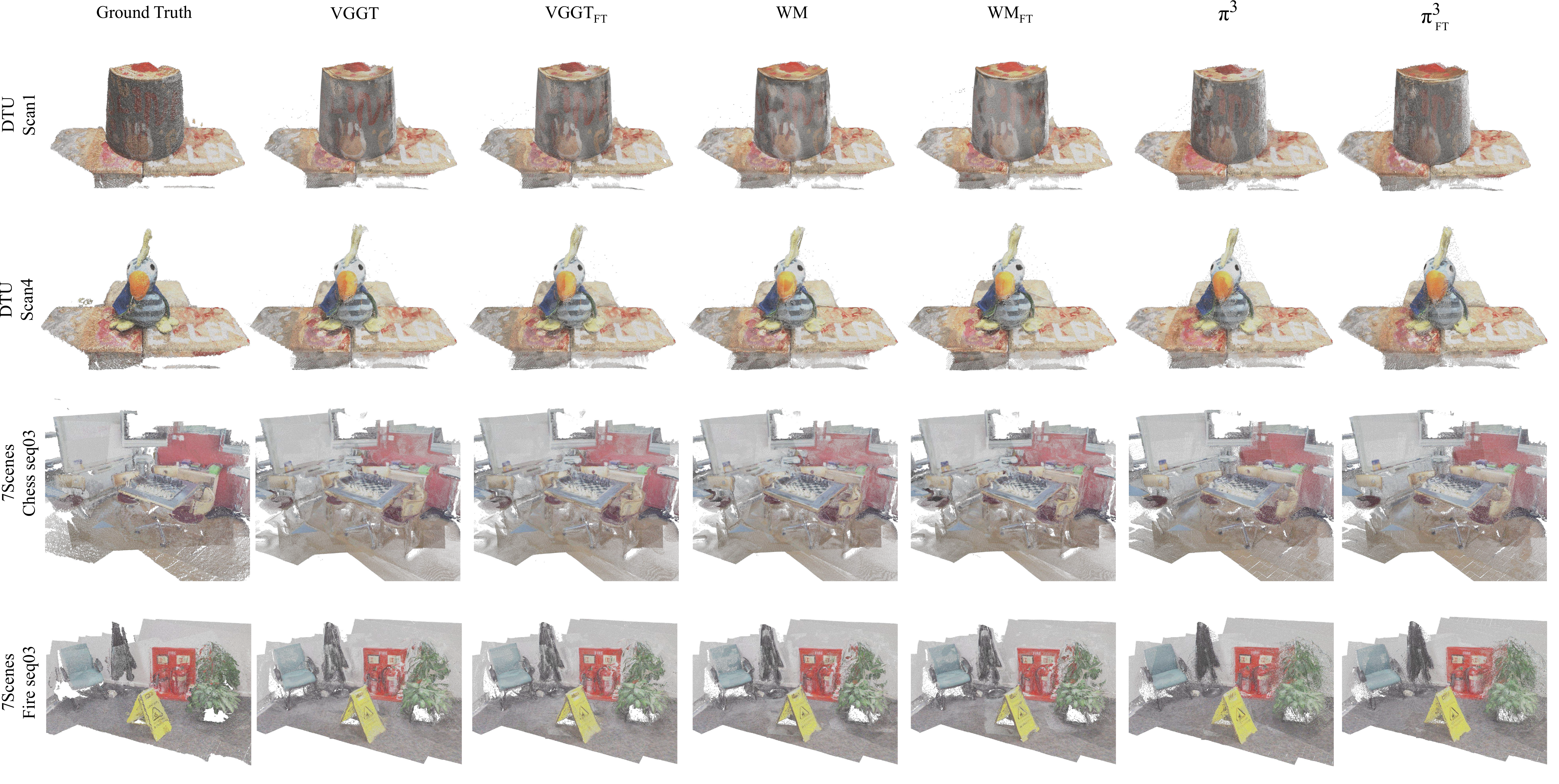}
    
    \caption{\textbf{DTU and 7Scenes Examples}. We show reconstruction results from the base and finetuned VGGT~\cite{wang2025vggt}, WorldMirror (WM)~\cite{liu2025worldmirroruniversal3dworld}, and \picubed~\cite{wang2025pi3} models on DTU's~\cite{jensen2014large} scan1 and scan4 and 7Scenes's~\cite{shotton2013scene} chess-seq03 and fire-seq03. The ground-truth reconstruction, obtained using Doppelgangers++~\cite{xiangli2025doppelgangersimprovedvisualdisambiguation} and MASt3R-SfM~\cite{duisterhof2025mastrsfm} as further detailed in the text, are shown in the first column. The predicted scenes are automatically aligned to the ground truth per the evaluation protocol discussed in the main paper.
    }
    \vspace{-0.5em}
    \label{fig:dtu_7scenes_vis_examples}
\end{figure*}

\parnobf{Additional Results: DTU and 7Scenes.}
We provide additional dense reconstruction experiments on the object-centric DTU~\cite{jensen2014large} and indoor 7Scenes~\cite{shotton2013scene} datasets. We follow \picubed~\cite{wang2025pi3} and sample every 5th image from DTU (10 images per scene). For 7Scenes, we sample every 200th image, corresponding to \picubed's dense-view evaluation setting (16 scenes with 25 images, 2 scenes with 13 images). We report accuracy (ACC) and completion (COMP) as in the main paper.

We depict results in Table ~\ref{tab:mv_recon_dtu_7scenes}. As shown on DTU, despite fine-tuning on Internet photos, all our finetuned models are able to generalize to an object-centric dataset: VGGT shows minimal change from fine-tuning, while WM and \picubed show minimal performance loss. Furthermore, on 7Scenes, the reconstruction performance for all models is extremely similar before and after fine-tuning. This is validated by the qualitative results in Figure~\ref{fig:dtu_7scenes_vis_examples}. Thus, our finetuned models still generalize to indoor scenes with minimal impact on dense image inputs.

\begin{figure*}[t]
    \centering
    \includegraphics[width=\linewidth]{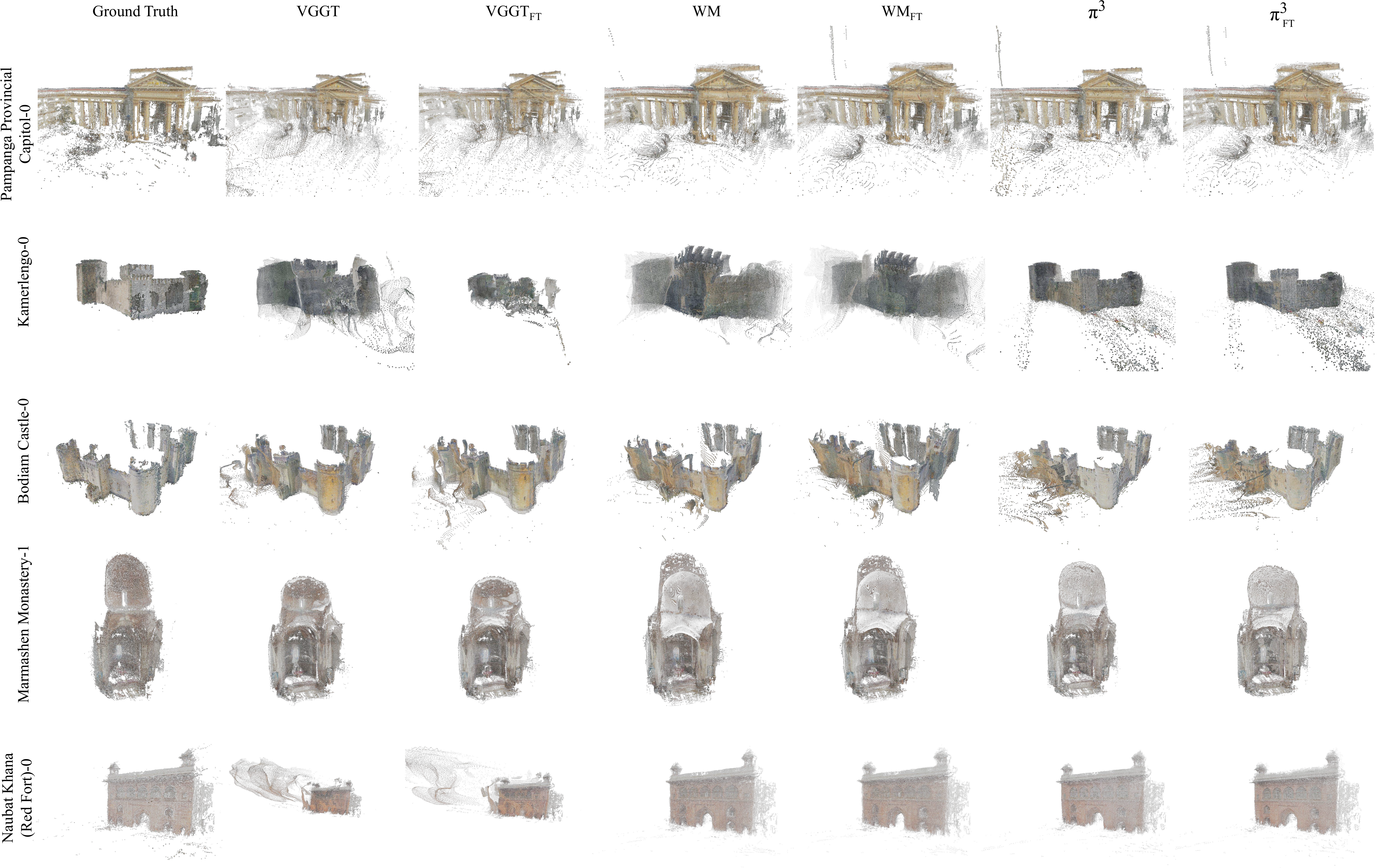}
    
    \caption{\textbf{Random UnSceneRecon Examples}. We show reconstruction results from the base and finetuned VGGT~\cite{wang2025vggt}, WorldMirror (WM)~\cite{liu2025worldmirroruniversal3dworld}, and \picubed~\cite{wang2025pi3} models on five randomly selected UnSceneRecon scenes. The selected scenes (top to bottom) are Pampanga Provincial Capitol-0, Kamerlengo-0, Bodiam Castle-0, Marmashen Monastery-1, and Naubat Khana (Red Fort)-0. The ground-truth reconstruction, obtained using Doppelgangers++~\cite{xiangli2025doppelgangersimprovedvisualdisambiguation} and MASt3R-SfM~\cite{duisterhof2025mastrsfm} as further detailed in the text, are shown in the first column. The predicted scenes are automatically aligned to the ground truth per the evaluation protocol discussed in the main paper.
    }
    \vspace{-0.7em}
    \label{fig:additional_unscenerecon_examples}
\end{figure*}

\parnobf{Additional Results: UnSceneRecon.} We show additional qualitative results on UnSceneRecon in Figure~\ref{fig:additional_unscenerecon_examples}; input images and 3D model visualizations for \picubed are shown in the accompanying \texttt{viewer.html}. These scenes are selected from the 100 reconstructions in UnSceneRecon using a random sampler. As shown, there is negligible difference in reconstruction quality between the base and finetuned models. Note that when the base model reconstructs poorly, the fine-tuned model does too, as exemplified by VGGT and WorldMirror on Kamerlengo-0 and Naubat Khana (Red Fort)-0. The scale alignment of VGGT\textsubscript{FT} appears much worse than base VGGT on Kamerlengo-0, since automatic alignment has greater variance when aligning incorrect point clouds to the ground truth. These reconstructions demonstrate the difficulty that current 3DFMs have in accurately reconstructing Internet photos, emphasizing the importance of a test set of real-world, unconstrained settings.

\subsection{Monocular Depth Estimation}
\label{sec:monodepth_eval}

We additionally evaluate monocular depth to determine if fine-tuning alters the performance of the 3DFM's dense prediction heads, as it directly alters the internal representations they decode. 

\parnobf{Experimental Details.} We test four datasets: Sintel~\cite{butler2012sintel}, Bonn~\cite{palazzolo2019bonn}, KITTI~\cite{geiger2013kitti}, and NYU-v2~\cite{silberman2012nyuv2}. Sintel and Bonn contain synthetic scenes; KITTI contains real outdoor driving scenes; NYU-v2 contains real indoor scenes. Following prior work~\cite{wang2025pi3, wang2025cut3r, zhang2024monst3r}, we align each predicted depth to the ground truth with per-frame median scaling. For VGGT and WorldMirror, we directly use the depth head outputs for evaluation. For \picubed, since the model does not have a depth head, we obtain the depth by taking the z-values of the model's point map prediction.

\parnobf{Metrics.} We report absolute relative error (AbsRel) and threshold accuracy below 1.25 ($\delta_1$) like in~\cite{wang2025pi3, wang2025cut3r, zhang2024monst3r}.

\parnobf{Results.} We show monocular depth results in Table~\ref{tab:monodepth}. Remarkably, all fine-tuned models perform similarly to their base counterparts across all datasets, with VGGT demonstrating minor improvements. This indicates that the frozen dense prediction heads remain effective at decoding the altered internal representations, despite fine-tuning with only rotation loss and no depth supervision.

\subsection{Extra Ablations}
\label{sec:supp-extra-ablations}

We show an expanded version of our ablation results in Table~\ref{tab:ablation_full_table}, with metrics for VGGT~\cite{wang2025vggt}, WorldMirror (WM)~\cite{liu2025worldmirroruniversal3dworld}, and \picubed~\cite{wang2025pi3}. For clarity, we show columns for whether models are trained on select layers only (\textbf{LO}) and bias only (\textbf{BO}); we additionally denote whether the reconstruction metrics use the point head or not (using fused point maps from unprojected depth) in the \textbf{PH} (point head) column. 
We use the same $\triangle$\textbf{REC}\textsubscript{PH} and $\triangle$\textbf{REC}\textsubscript{Fused} metrics as in the main paper's ablation table. 
We show the same rotation metrics: \textbf{MRE},
\textbf{RA}\textsubscript{15}, and \textbf{RA}\textsubscript{30},
as well as the median reconstruction metrics: \textbf{ACC} and \textbf{COMP}, as discussed in the main paper. \textbf{REC} is the average of \textbf{ACC} and \textbf{COMP}.

As discussed in the main paper, \picubed performs similarly to WorldMirror (WM) in the ablations: regarding how to fine-tune the backbone, using select-layers only and bias-only provides a good trade-off between rotation and reconstruction performance (-26.8 for $\triangle$\textbf{ROT} and 9.2 for $\triangle$\textbf{REC}). Switching the former option to \textit{all layers} or the latter option to \textit{weights and biases} leads to a performance degradation in \textbf{REC} (from a 9.2 $\triangle$\textbf{ROT} to 14.0 and 15.8, respectively). Interestingly, our fine-tuned \picubed model on all layers and weights and biases exhibits abnormally strong performance (a $\triangle$\textbf{ROT} of -37.7 and $\triangle$\textbf{REC} of 9.9), and is an outlier in the trends we see across all three models.

We also show \picubed's fine-tuning results when unfreezing only the camera head $\mathcal{D}_c$. Unlike VGGT~\cite{wang2025vggt} and WM~\cite{liu2025worldmirroruniversal3dworld}, which directly infer 3D points in global space, \picubed uses the extrinsic predictions of the camera head to transform predicted points in local coordinates to global coordinates. Consequently, we see that fine-tuning the camera head leads to worse $\triangle$\textbf{REC} metrics (567.4) compared other fine-tuning schemes; this indicates that the performance of $\mathcal{D}_c$ is destroyed. At the same time, we see that $\mathcal{D}_c$ does not have much capacity to align to our target task of extreme rotation estimation, reflected by a negligible $\triangle$\textbf{ROT} of 4.1.

\newpage

\begin{table*}[t]
    \centering
    \begin{minipage}[t]{0.48\linewidth}
    \vspace{-30mm}
        \captionof{table}{Evaluation of fine-tuned VGGT, WM, and \picubed models trained with additional translation loss (TL) on the \wELPt{} test set. Comparing with our final fine-tuned checkpoints (FT), the results show that translation supervision offers no improvement in relative translation and rotation accuracy.}
\setlength{\tabcolsep}{2pt}
\def\arraystretch{1}
\centering
\resizebox{\columnwidth}{!}{%
\begin{tabular}{llcccccc}
\toprule
\multicolumn{2}{c}{} &
\multicolumn{6}{c}{UnScenePairs-t} \\
\cline{3-8}
& Method &
MRE & RA$_{15}$ & RA$_{30}$ &
MTE & TA$_{15}$ & TA$_{30}$ \\
\midrule

\multirow{6}{*}{\rotatebox[origin=c]{90}{Large}} &
 \vggtft &
1.08 & 99.9 & 99.9 & 2.24 & 94.0 & 97.6 \\

& \vggttrans &
1.15 & 99.8 & 99.8 & 3.08 & 93.0 & 96.9 \\

& \worldmirrorft &
1.15 & 99.4 & 99.7 & 3.34 & 88.7 & 94.6 \\

& \worldmirrortrans &
1.22 & 99.4 & 99.7 & 3.85 & 88.0 & 94.2 \\

& \picubedft &
0.99 & 99.9 & 100.0 & 3.05 & 88.7 & 95.9 \\

& \picubedtrans &
1.52 & 99.8 & 100.0 & 6.93 & 80.6 & 92.8 \\
\midrule

\multirow{6}{*}{\rotatebox[origin=c]{90}{Small}} &
\vggtft &
2.04 & 100.0 & 100.0 & 9.18 & 61.4 & 75.1 \\

& \vggttrans &
2.36 & 99.8 & 100.0 & 10.34 & 59.1 & 73.6 \\

& \worldmirrorft &
2.39 & 97.9 & 99.6 & 11.75 & 56.4 & 72.1 \\

& \worldmirrortrans &
2.43 & 97.3 & 99.8 & 12.47 & 54.3 & 72.1 \\

& \picubedft &
1.99 & 99.0 & 100.0 & 10.91 & 57.5 & 75.5 \\

& \picubedtrans &
2.38 & 99.0 & 100.0 & 14.71 & 51.8 & 70.0 \\
\midrule

\multirow{6}{*}{\rotatebox[origin=c]{90}{None}} &
\vggtft &
14.48 & 50.6 & 62.1 & 35.79 & 26.5 & 44.0 \\

& \vggttrans &
14.75 & 50.2 & 61.6 & 37.75 & 22.2 & 39.5 \\

& \worldmirrorft &
13.13 & 53.3 & 64.5 & 33.42 & 27.8 & 46.7 \\

& \worldmirrortrans &
13.36 & 52.0 & 63.0 & 38.50 & 23.2 & 40.2 \\

& \picubedft &
13.31 & 53.1 & 65.5 & 32.05 & 31.4 & 47.7 \\

& \picubedtrans &
15.50 & 49.0 & 63.8 & 33.24 & 27.8 & 46.0 \\

\bottomrule
\end{tabular}
}
\label{tab:og_loss}
    \end{minipage}\hfill
\begin{minipage}[t]{0.48\linewidth}
    \begin{minipage}{\linewidth}
    \centering
        \captionof{table}{
        \textbf{Monocular Depth Estimation} on Sintel~\cite{butler2012sintel}, Bonn~\cite{palazzolo2019bonn}, KITTI~\cite{geiger2013kitti}, and NYU-v2~\cite{silberman2012nyuv2} datasets. Non-negligible differences ($>5\%$ relative) between the base and fine-tuned models are in \textbf{bold}.
    }
    \tablestyle{1pt}{1.0}
    \resizebox{\columnwidth}!{
        \begin{tabular}{@{}lcccccccc@{}}
            \toprule
            \multirow{2}{*}{\textbf{Method}} &
            \multicolumn{2}{c}{\textbf{Sintel}} &
            \multicolumn{2}{c}{\textbf{Bonn}} &
            \multicolumn{2}{c}{\textbf{KITTI}} &
            \multicolumn{2}{c}{\textbf{NYU-v2}} \\
            \cmidrule(r){2-3} \cmidrule(r){4-5} \cmidrule(r){6-7} \cmidrule(r){8-9}
            &
            AbsRel$\downarrow$ & $\delta_1\!\uparrow$ &
            AbsRel$\downarrow$ & $\delta_1\!\uparrow$ &
            AbsRel$\downarrow$ & $\delta_1\!\uparrow$ &
            AbsRel$\downarrow$ & $\delta_1\!\uparrow$ \\
            \midrule

VGGT~\cite{wang2025vggt} & 0.335 & 0.597 & 0.053 & 0.970 & 0.082 & 0.947 & 0.056 & 0.951 \\
\vggtft & \textbf{0.316} & 0.621 & \textbf{0.050} & 0.974 & \textbf{0.077} & 0.952 & \textbf{0.052} & 0.955 \\

            \midrule

WM~\cite{liu2025worldmirroruniversal3dworld} & 0.340 & 0.625 & 0.066 & 0.963 & 0.093 & 0.930 & 0.053 & 0.957 \\
\worldmirrorft & 0.336 & 0.633 & 0.064 & 0.964 & 0.089 & 0.933 & 0.053 & 0.957 \\

            \midrule

\picubed~\cite{wang2025pi3} & 0.280 & 0.617 & 0.048 & 0.974 & 0.059 & 0.971 & 0.054 & 0.956 \\
\picubedft & 0.287 & 0.596 & 0.048 & 0.974 & 0.061 & 0.969 & 0.054 & 0.956 \\

            \bottomrule
        \end{tabular}
    }
    \label{tab:monodepth}
    \end{minipage}

    \vspace{0.9cm} %

    \begin{minipage}{\linewidth}
    \centering
        \captionof{table}{
        \textbf{Dense Reconstruction} on DTU~\cite{jensen2014large} and 7Scenes~\cite{shotton2013scene} datasets. Non-negligible differences ($>5\%$ relative) between the base and fine-tuned models are in \textbf{bold}.
    }
    \vspace{-0.75em}
    \tablestyle{3pt}{1.0}
    \resizebox{\columnwidth}!{
    \begin{tabular}{l cccc cccc}
        \toprule
        {\multirow{4}{*}{\textbf{Method}}} &
        \multicolumn{4}{c}{\textbf{DTU}} &
        \multicolumn{4}{c}{\textbf{7Scenes}} \\
        \cmidrule(r){2-5} \cmidrule(r){6-9}
        &
        \multicolumn{2}{c}{ACC $\downarrow$}  &
        \multicolumn{2}{c}{CMP $\downarrow$} &
        \multicolumn{2}{c}{ACC $\downarrow$}  &
        \multicolumn{2}{c}{CMP $\downarrow$} \\
        \cmidrule(r){2-3} \cmidrule(r){4-5} \cmidrule(r){6-7} \cmidrule(r){8-9}
        &
        Mean & Med. &
        Mean & Med. &
        Mean & Med. &
        Mean & Med. \\
        \midrule

VGGT~\cite{wang2025vggt} & 1.185 & 0.716 & 2.215 & 1.302 & 0.020 & 0.007 & 0.030 & 0.014 \\
\vggtft & 1.178 & 0.711 & 2.189 & 1.256 & \textbf{0.018} & 0.007 & 0.029 & 0.014 \\
        
        \midrule
        
WM~\cite{liu2025worldmirroruniversal3dworld} & \textbf{1.033} & \textbf{0.573} & 1.759 & 0.790 & \textbf{0.016} & 0.007 & 0.028 & 0.013 \\
\worldmirrorft & 1.121 & 0.614 & 1.761 & 0.783 & 0.017 & 0.007 & 0.027 & \textbf{0.012} \\

        \midrule

\picubed~\cite{wang2025pi3} & \textbf{1.152} & \textbf{0.622} & 1.797 & 0.631 & 0.016 & 0.007 & 0.022 & 0.011 \\
\picubedft & 1.396 & 0.748 & 1.768 & 0.634 & 0.016 & 0.007 & \textbf{0.020} & \textbf{0.008} \\

        \bottomrule
    \end{tabular}
    }
    \label{tab:mv_recon_dtu_7scenes}
    \end{minipage}
\end{minipage}

    \vspace{0.5cm} %

    \begin{minipage}{\textwidth}
    \centering
    \captionof{table}{
        \textbf{Full Ablation Table.} The full table for the ablation evaluating rotation ($\triangle$\textbf{ROT}) and reconstruction
        ($\triangle$\textbf{REC}) changes relative to pretrained models. The camera head is denoted as $\mathcal{D}_c$ and the backbone as AA. 
        We show all fine-tuned results for layer-only (LO), bias-only (BO), or both (LO+BO) updates.
        We denote whether we use the point head or not (using fused unprojected depths) in the \textbf{PH} column for reconstruction metrics. \textbf{REC} is the average of \textbf{COMP} and \textbf{ACC}; we report median values only.
        The reconstruction delta $\triangle$\textbf{REC} is \textbf{REC}'s percent change compared to the base model.
        Significant improvements ($>10\%$ error drop) for $\triangle$\textbf{ROT} and $\triangle$\textbf{REC} are shown in
        \textcolor{ForestGreen}{green}, and degradations in \textcolor{BrickRed}{red}.
    }
    \vspace{-2mm}
    \tablestyle{1pt}{1}
    \scriptsize
    \setlength{\tabcolsep}{3pt}
    \renewcommand{\arraystretch}{0.8}

    \newcolumntype{C}{>{\centering\arraybackslash}p{0.8em}}

    \resizebox{\linewidth}!{
    \begin{tabular}{lcccc |@{\hspace{0.6em}} c ccc @{\hspace{0.6em}}| cc ccc @{\hspace{0.6em}}|c}
        \toprule
        
        {\multirow{0.9}{*}{\textbf{Model}}} &
        {\multirow{0.9}{*}{\textbf{Comp.}}} &
        {\multirow{0.9}{*}{\hspace{-0.3em}\textbf{LO}}} &
        {\multirow{0.9}{*}{\hspace{-0.3em}\textbf{BO}}} &
        {\multirow{0.9}{*}{\hspace{-0.2em}\textbf{PH}}} &
        {\multirow{0.9}{*}{$\triangle\textbf{ROT}$}} &
        {\multirow{0.9}{*}{\textbf{MRE}}} &
        {\multirow{0.9}{*}{\textbf{RA\textsubscript{15}}}} &
        {\multirow{0.9}{*}{\textbf{RA\textsubscript{30}}}} &
        {\multirow{0.9}{*}{$\triangle\textbf{REC}_{\textbf{PH}}$}} &
        {\multirow{0.9}{*}{$\triangle\textbf{REC}_{\textbf{Fused}}$}} &
        {\multirow{0.9}{*}{\textbf{REC}}} &
        {\multirow{0.9}{*}{\textbf{ACC}}} &
        {\multirow{0.9}{*}{\textbf{COMP}}} &
        {\multirow{0.9}{*}{\textbf{\#Params}}} \\
        \midrule
        
        \multirow{7}{*}{\centering VGGT}
          & $\mathcal{D}_c$      & × & × & \checkmark
          & 6.8 & 33.79 & 28.2  & 46.4
          & 0.0 & (N/A) & 0.889 & 1.049 & 0.729
          & 216.2M \\
          & AA+$\mathcal{D}_c$   & × & × & ×
          & \textcolor{ForestGreen}{-74.3} & 8.14  & 65.5 & 78.5
          & (N/A) & \textcolor{BrickRed}{90.3} & 1.692 & 1.384 & 2.000
          & 820.9M \\
          & AA      & × & × & ×
          & \textcolor{ForestGreen}{-69.8} & 9.57  & 60.4 & 73.9
          & (N/A) & -9.2 & 0.808 & 0.961 & 0.654
          & 604.7M \\
          & AA      & × & × & \checkmark &
          \textcolor{ForestGreen}{-69.8} & 9.57  & 60.4 & 73.9
          & \textcolor{ForestGreen}{-33.7} & (N/A) & 0.590 & 0.687 & 0.493
          & 604.7M \\
          & AA   & \checkmark & × & \checkmark
          & \textcolor{ForestGreen}{-66.7} & 10.55 & 57.5 & 70.8
          & \textcolor{ForestGreen}{-33.3} & (N/A) & 0.593 & 0.727 & 0.459
          & 100.8M \\
          & AA      & × & \checkmark & \checkmark
          & \textcolor{ForestGreen}{-69.7} & 9.60   & 61.3 & 75.6
          & \textcolor{ForestGreen}{-16.7} & (N/A) & 0.741 & 0.912  & 0.570
          & 0.4M \\
          & AA   & \checkmark & \checkmark & \checkmark
          & \textcolor{ForestGreen}{-59.8} & 12.71 & 53.6  & 67.9
          & \textcolor{ForestGreen}{-12.4} & (N/A) & 0.779 & 0.908 & 0.650
          & 0.07M \\
        
        \midrule
        
        \multirow{7}{*}{\centering \hspace{0.5em}WM}
        & $\mathcal{D}_c$      & × & × & \checkmark
        & -1.4 & 18.98 & 43.8  & 60.7
        & 0.0 & (N/A) & 0.500 & 0.612 & 0.387
        & 216.2M \\
        & AA+$\mathcal{D}_c$   & × & × & ×
        & \textcolor{ForestGreen}{-47.5} & 10.10  & 60.7 & 75.8
        & (N/A) & \textcolor{BrickRed}{81.5} & 0.907 & 1.087 & 0.727
        & 820.9M \\
        & AA      & × & × & ×
        & \textcolor{ForestGreen}{-41.6} & 11.24 & 58.0 & 71.3
        & (N/A) & \textcolor{BrickRed}{13.6} & 0.567 & 0.704 & 0.431
        & 604.7M \\
        & AA      & × & × & \checkmark
        & \textcolor{ForestGreen}{-41.6} & 11.24 & 58.0 & 71.3
        & \textcolor{BrickRed}{14.0} & (N/A) & 0.570 & 0.716 & 0.423
        & 604.7M \\
        & AA   & \checkmark & × & \checkmark
        & \textcolor{ForestGreen}{-40.4} & 11.48 & 55.9 & 70.0
        & \textcolor{BrickRed}{18.3} & (N/A) & 0.591 & 0.719 & 0.463
        & 100.8M \\
        & AA      & × & \checkmark & \checkmark
        & \textcolor{ForestGreen}{-42.3} & 11.11 & 56.9 & 70.6
        & \textcolor{BrickRed}{12.9} & (N/A) & 0.564 & 0.702 & 0.426
        & 0.4M \\
        & AA   & \checkmark & \checkmark & \checkmark
        & \textcolor{ForestGreen}{-39.0} & 11.75 & 56.2  & 68.1
        & 2.9 & (N/A) & 0.514 & 0.660 & 0.368
        & 0.07M \\
        
        \midrule
        
        \multirow{6}{*}{\centering \hspace{1em}\picubed}
        & $\mathcal{D}_c$      & × & × & \checkmark
        & 4.1 & 18.39 & 44.0    & 59.3
        & \textcolor{BrickRed}{567.4} & (N/A) & 2.812 & 2.755 & 2.869
        & 2.1M \\

        & AA      & × & × & \checkmark &
        \textcolor{ForestGreen}{-37.7} & 11.00    & 59.5 & 73.0
        & 9.9 & (N/A) & 0.463 & 0.501 & 0.425
        & 453.5M \\
        & AA   & \checkmark & × & \checkmark
        & \textcolor{ForestGreen}{-31.3} & 12.13 & 55.0 & 69.8
        & \textcolor{BrickRed}{14.0} & (N/A) & 0.480 & 0.562 & 0.398
        & 113.4M \\
        & AA      & × & \checkmark & \checkmark
        & \textcolor{ForestGreen}{-39.6} & 10.66 & 60.2 & 73.8
        & \textcolor{BrickRed}{15.8} & (N/A) & 0.488 & 0.555 & 0.421
        & 0.3M \\
        & AA   & \checkmark & \checkmark & \checkmark
        & \textcolor{ForestGreen}{-26.8} & 12.92 & 54.0    & 69.2
        & 9.2 & (N/A) & 0.460 & 0.517 & 0.403
        & 0.08M \\

        \bottomrule
    \end{tabular}
    }
    \label{tab:ablation_full_table}
    \end{minipage}

\end{table*}

\end{document}